\documentclass[letterpaper]{article} 
\usepackage[]{aaai25}
\nocopyright
\usepackage{times}  
\usepackage{helvet}  
\usepackage{courier}  
\usepackage[hyphens]{url}  
\usepackage{enumitem}
\usepackage{graphicx} 
\usepackage{natbib}  
\usepackage{caption} 
\usepackage{bm,amssymb,amsmath}
\usepackage{algorithm}
\usepackage{algorithmic}
\newtheorem{lemma}{Lemma}
\usepackage{color}

\usepackage{newfloat}
\usepackage{listings}
\usepackage{booktabs}
\usepackage{float}
\usepackage{subfig}
\DeclareCaptionStyle{ruled}{labelfont=normalfont,labelsep=colon,strut=off} 
\floatstyle{ruled}
\newfloat{listing}{tb}{lst}{}
\floatname{listing}{Listing}
\def\method{NeGPR}
\title{Nested Graph Pseudo-Label Refinement for Noisy Label\\ Domain Adaptation Learning}
\author{
    Yingxu Wang\textsuperscript{\rm 1},
    Mengzhu Wang\textsuperscript{\rm 2},
    Zhichao Huang\textsuperscript{\rm 3},
    Suyu Liu\textsuperscript{\rm 4},
    Nan Yin\textsuperscript{\rm 5}
}
\affiliations{
    \textsuperscript{\rm 1}Mohamed bin Zayed University of Artificial Intelligence \quad
    \textsuperscript{\rm 2}Hebei University of Technology\\
    \textsuperscript{\rm 3}JD Industrial\quad
    \textsuperscript{\rm 4}Nanyang Technological University \\
    \textsuperscript{\rm 5}Hong Kong University of Science and Technology \\

    yingxv.wang@gmail.com, \quad dreamkily@gmail.com, \quad iceshzc@gmail.com, \\ \quad suyu.liu@ntu.edu.sg,
    \quad yinnan8911@gmail.com
    
}

\usepackage{bibentry}

\begin{document}

\maketitle

\begin{abstract}

Graph Domain Adaptation (GDA) facilitates knowledge transfer from labeled source graphs to unlabeled target graphs by learning domain-invariant representations, which is essential in applications such as molecular property prediction and social network analysis. However, most existing GDA methods rely on the assumption of clean source labels, which rarely holds in real-world scenarios where annotation noise is pervasive. 
This label noise severely impairs feature alignment and degrades adaptation performance under domain shifts. To address this challenge, we propose \textbf{Ne}sted \textbf{G}raph \textbf{P}seudo-Label \textbf{R}efinement (\method{}), a novel framework tailored for graph-level domain adaptation with noisy labels. \method{} first pretrains dual branches, i.e., semantic and topology branches, by enforcing neighborhood consistency in the feature space, thereby reducing the influence of noisy supervision. To bridge domain gaps, \method{} employs a nested refinement mechanism in which one branch selects high-confidence target samples to guide the adaptation of the other, enabling progressive cross-domain learning. Furthermore, since pseudo-labels may still contain noise and the pre-trained branches are already overfitted to the noisy labels in the source domain, \method{} incorporates a noise-aware regularization strategy. This regularization is theoretically proven to mitigate the adverse effects of pseudo-label noise, even under the presence of source overfitting, thus enhancing the robustness of the adaptation process. Extensive experiments on benchmark datasets demonstrate that \method{} consistently outperforms state-of-the-art methods under severe label noise, achieving gains of up to 12.7\% in accuracy.

\end{abstract}
\section{Introduction}


Graph Domain Adaptation (GDA)~\cite{you2022bringing,cai2024graph} has emerged as a prominent technique for leveraging labeled graph data from a source domain to enhance learning on an unlabeled target graph domain. Its efficacy has been demonstrated across diverse applications, including temporally-evolved social network analysis~\citep{wang2021inductive}, molecular property prediction~\citep{zhu2023explaining}, and protein-protein interaction modeling~\citep{cho2016compact}. The core paradigm typically involves learning domain-invariant node/graph representations that bridge the distributional shift between source and target domains, thus enabling effective inference on the target data.

However, the success of standard GDA methods crucially relies on the accurately labeled source data. In practice, source domain labels are often corrupted by noise arising from annotation errors~\citep{dai2021nrgnn,yuan2023learning}, subjective judgments~\citep{platanios2016estimating}, or inherent ambiguities in data collection~\citep{chen2020negative}. This prevalent issue of label noise can severely misguide the learning of domain-invariant representations~\citep{li2020dividemix}, leading to suboptimal or even detrimental adaptation performance on the target domain~\citep{yin2025roda}. 
Existing noise label learning methods typically rely on loss function design to mitigate the impact of noisy labels~\citep{han2018co,zhang2021robust}, which selects clean instances for joint training, and robust loss functions~\citep{wei2020combating,li2020dividemix}, which leverage small-loss selection or instance mixture models. While effective in controlled settings, these approaches fall short in the presence of domain shifts. The coexistence of distribution shift and label noise leads to misaligned feature spaces, causing noise-robust losses to erroneously align clean features with noisy targets, thereby amplifying negative transfer~\citep{yu2020label}.
While recent efforts have been made to address GDA under noisy labels~\cite{yuan2023alex,wang2022bayesian}, they primarily target node classification tasks, leaving a critical gap in addressing graph-level scenarios. Many real-world applications, such as molecular property prediction~\citep{stokes2020deep} and social network analysis~\citep{hamilton2017inductive}, inherently depend on graph-level classification, where label noise can severely compromise the identification of functional groups and the modeling of community behaviors. The lack of attention to graph-level adaptation under noisy labels significantly limits the practical applicability of existing methods in high-impact domains.

In this paper, we investigate the development of an efficient GDA framework for scenarios involving label noise. However, designing such a framework poses several fundamental challenges: (1) \textit{Distribution shift undermines loss-based denoising.} 
Conventional noise-robust loss functions are primarily designed for specific domains and often struggle under distribution shifts. In the presence of noisy labels in the source domain, aligning target features with corrupted source representations can lead to noise-aligned embeddings, degrading generalization due to feature misalignment and increased risk of negative transfer. Recent studies in GDA have highlighted that noisy supervision severely hinders feature alignment across domains, especially when relying on pseudo labels or unreliable source signals~\citep{yuan2023alex}. These findings underscore the need for noise-aware mechanisms that explicitly account for both label noise and domain discrepancy. 
(2) \textit{Imperfect pseudo labels compromise domain adaptation.} Probability-based pseudo-labeling has shown promise in bridging distribution shift and mitigating supervision noise~\citep{yuan2023alex,yin2023coco}. However, the reliability of selected pseudo labels is often compromised by erroneous source annotations, leading to residual noise being transferred into the target domain. In Graph Neural Networks (GNNs), such corrupted pseudo labels can propagate through message passing, triggering self-reinforcing error cascades. As each GNN layer aggregates information from potentially mislabeled neighbors, the accumulated noise progressively deteriorates local neighborhood structures and distorts global representations over successive adaptation rounds~\citep{wang2024noisygl}.
(3) \textit{Label noise impairs distribution alignment in GDA.}
Existing methods typically adopt explicit~\citep{long2015learning} or implicit~\citep{long2018conditional} strategies to align feature distributions across domains. However, significant label noise corrupts supervision signals, causing samples to drift toward incorrect class regions and disrupting the formation of domain-invariant features. This misalignment undermines the effectiveness of domain discriminators and hampers reliable adaptation. These challenges call for a unified framework that combines noise-robust representation learning, trustworthy pseudo-label refinement, and alignment strategies that preserve class-level semantics across domains.

To tackle these challenges, we propose \textbf{Ne}sted \textbf{G}raph \textbf{P}seudo-Label \textbf{R}efinement (\method{}), a novel framework designed for GDA under noisy labels.
To effectively disentangle the impact of label noise from domain distribution shift, \method{} first pre-trains noise-resilient models from implicit and explicit perspectives by enforcing semantic consistency among neighboring samples in the feature space.
The implicit branch promotes feature-level consistency based on learned representations, while the explicit branch captures structural patterns by leveraging graph topology. This dual-perspective design improves robustness to noisy supervision and provides a reliable foundation for domain adaptation.
Then, to align the domain distribution, \method{} iteratively leverages cross-branch knowledge, where one branch filters highly reliable target domain samples, and the other branch is fine-tuned accordingly, enabling mutual enhancement and progressive adaptation. However, the filtered pseudo-labels may still contain erroneous category information, and the pre-trained branches have already overfitted to the label noise in the source domain. The interplay of these two factors exacerbates performance degradation during domain adaptation. To tackle this, \method{} employs a regularization along with a theoretical analysis demonstrating its effectiveness in suppressing the influence of noisy pseudo-labels. Extensive experiments demonstrate that \method{} significantly outperforms state-of-the-art methods under severe label noise.
Our main contributions are summarized as: 
\begin{itemize}[itemsep=2pt,topsep=0pt,parsep=0pt]
    \item We investigate a novel problem setting, graph domain adaptation learning under label noise, where label noise and domain shift coexist and jointly pose significant challenges for graph representation learning. 
    \item We propose \method{}, a dual-branch framework that integrates noise-resilient pre-training, nested pseudo-label refinement, and theoretically grounded regularization to tackle graph domain adaptation under label noise. 
    \item We evaluate \method{} on extensive datasets, showing that \method{} significantly outperforms existing approaches under various noise levels and domain shift scenarios.
\end{itemize}
\section{Related work}

\paragraph{Graph Domain Adaptation.} Graph Domain Adaptation (GDA) has emerged as a critical research topic, aiming to leverage labeled source domain graphs to enable robust representation learning on unlabeled or sparsely labeled target graphs \cite{lin2023multi,luo2023source,liu2024rethinking}. To achieve this, most existing approaches first employ Graph Neural Networks (GNNs)~\citep{kipf2017semi} to generate representations for each graph \cite{wu2022attraction,zhu2021transfer, yin2022deal}. They then commonly use adversarial learning to implicitly align feature distributions and reduce domain discrepancies, apply pseudo-labeling to iteratively refine predictions in the target domain, or incorporate structure-aware strategies to explicitly align graph-level semantics and topological structures, thereby improving generalization across diverse graph domains \cite{yin2023coco, wang2024degree,liu2024pairwise}. However, these methods often overlook the impact of noisy labels, which can distort learned representations and lead to misaligned distributions and unreliable predictions in the target domain. Although a few label-denoising GDA methods have been proposed, they primarily focus on node-level tasks \cite{yuan2023alex}. To address these limitations, we propose a novel label-denoising domain adaptation method designed for graph-level classification tasks.

\begin{figure*}[t!]
  \centering
  \includegraphics[width=0.95\textwidth]{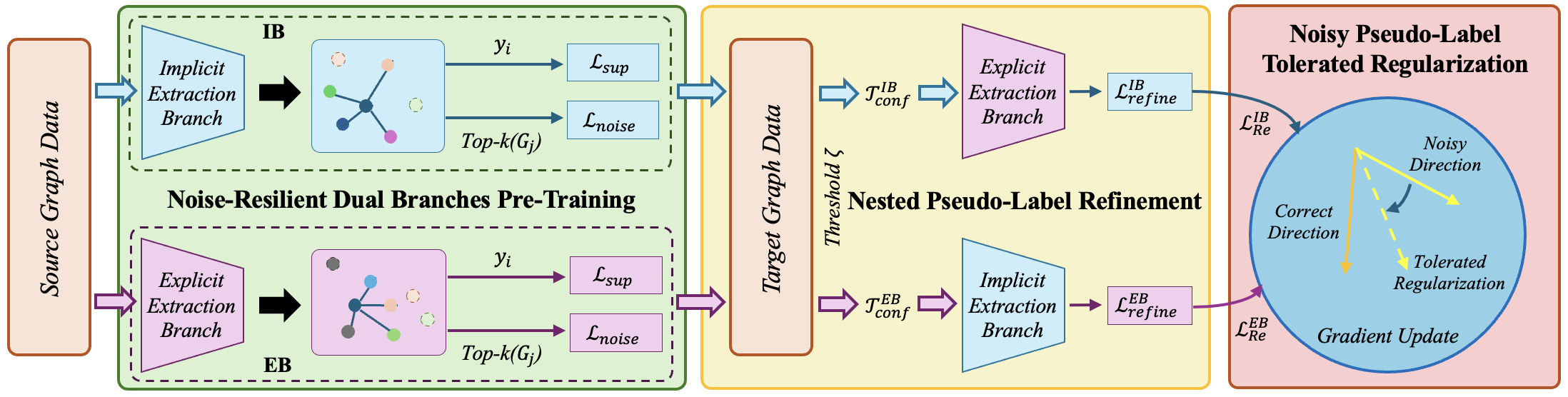} 
\caption{Overview of the proposed \method{}. \method{} consists of a dual-branch pretraining module that captures complementary semantic and structural features under label noise. Then, a nested pseudo-label refinement module alternately selects high-confidence target samples from one branch to guide the other, enabling progressive cross-domain adaptation. The noisy pseudo-label tolerated regularizatio penalizes overconfident predictions to suppress the effect of noisy pseudo labels.}
  \label{framework}
\end{figure*}

\noindent\textbf{Learning with Noise Labels.} Learning with noisy labels has garnered significant attention for its crucial role in developing robust models under imperfect supervision, which has been widely used in machine learning and computer vision  \cite{zhu2024robust}. Existing methods typically address label noise by employing robust loss functions, identifying and filtering out noisy samples, or refining labels through correction mechanisms \cite{feng2021can, xu2025unraveling}. However, existing methods still insufficiently investigate the interplay between label noise and domain adaptation \cite{yin2024sport, zhu2024robust}. In particular, applying a model trained on the source domain to the target domain can be regarded as a noisy inference process due to distributional shifts inherent in domain adaptation \cite{yu2020label, dan2024tfgda}. Furthermore, label noise in the source domain can also degrade model performance \cite{yuan2023alex, yu2024curriculum}. Critically, conventional methods cannot disentangle whether the observed performance degradation is primarily attributable to domain shift or label noise, thereby limiting their ability to address the underlying causes of adaptation failure effectively. To address this challenge, we propose a novel learning framework designed to mitigate the effects of domain shift and label noise.  

\section{Methodology}

\subsection{Overview of Framework}
This work studies the problem of unsupervised graph domain adaptation in the presence of noisy labels and proposes a novel framework, \method{}, as illustrated in Fig.~\ref{framework}.
\method{} comprises three key components: (1) \textbf{Noise-Resilient Dual Branches Pre-Training.} To effectively suppress the impact of label noise, we first pre-train noise-resilient models from implicit and explicit perspectives by enforcing semantic consistency among neighboring samples in the feature space; 
(2) \textbf{Nested Pseudo-Label Refinement.} To align category-level distributions, each branch selects high-confidence pseudo-labeled target samples based on prediction confidence and uses them to fine-tune the other branch. This cross-branch refinement mitigates error accumulation from noisy pseudo labels and enables progressive domain adaptation through mutual supervision;
(3) \textbf{Noisy Pseudo-Label Tolerated Regularization.} To alleviate the negative impact of noisy pseudo labels, we introduce a noise-aware regularization term with theoretical guarantees. This regularization effectively suppresses error propagation induced by noisy pseudo labels during the adaptation process.

\subsection{Problem Formulation}
Given a graph $G = (\mathcal{V}, \mathcal{E}, \mathbf{X})$ with the set of nodes $\mathcal{V}$ and edges $\mathcal{E} \subseteq \mathcal{V} \times \mathcal{V}$. The $\mathbf{X} \in \mathbb{R}^{|\mathcal{V}| \times d}$ is the node feature matrix, where each row $\mathbf{x}_v \in \mathbb{R}^{d}$ denotes the feature of node $v \in \mathcal{V}$, $|\mathcal{V}|$ is the number of nodes, and $d$ denotes the dimension of node features.
In our setting, we have access to a labeled source domain $\mathcal{D}^s = \{(G_i^s, y_i^s)\}_{i=1}^{n_s}$ with $n_s$ samples, where the labels $y_i^s$ may be corrupted by noise, and an unlabeled target domain $\mathcal{D}^t = \{G_j^t\}_{j=1}^{n_t}$ with $n_t$ samples. Both domains share the label space $\mathcal{Y} =\{1,2,\cdots, C\}$ but follow different data distributions. The goal is to train the graph classification model using both $\mathcal{D}^s$ and $\mathcal{D}^t$ and achieve high accuracy on the target domain.

\subsection{Noise-Resilient Dual Branches Pre-Training}
To mitigate the adverse impact of noisy labels in the source domain, we adopt a dual-branch architecture that captures semantic consistency from implicit and explicit perspectives. Noisy supervision can distort the feature space by pulling semantically similar graphs toward incorrect class boundaries. In contrast, the local relationships among neighboring samples often remain reliable despite label corruption. Motivated by this, we construct two parallel branches that exploit neighborhood consistency to learn robust representations. One branch captures semantic similarity through learned features, while the other incorporates structural information derived from graph topology. This design enhances the model’s resilience to noise and provides a stable foundation for subsequent domain adaptation.

\noindent\textbf{Implicit Extraction Branch.} The implicit branch follows the MPNNs mechanism~\cite{gilmer2017neural}, which extracts graph semantics by aggregating neighborhood representations to update the central node embeddings. Specifically, we update the embedding of node $u$ at layer $l$ and then summarize the node embeddings into graph-level: 
\begin{equation}
\begin{aligned}
\label{eq:readout}
\mathbf{h}_{u}^{l}= &\operatorname{COM}\left(\mathbf{h}_{u}^{l-1}, \operatorname{AGG} \left(\mathbf{h}_{v\in \mathcal{N}(u)}^{l-1} \right)\right),
\\ \mathbf{z}_{G}^{IB}&=\operatorname{READOUT}\left(\left\{\mathbf{h}_{u}^{L}\right\}_{u \in \mathcal{V}}\right),
\nonumber
\end{aligned}
\end{equation}
where $\mathcal{N}(u)$ is the neighbours of node $u$. $\operatorname{COM}$ and $\operatorname{AGG}$ denote the combination and aggregation operations, $\operatorname{READOUT}$ is the graph pooling function. 
This formulation allows the implicit branch to capture structural information indirectly through supervised learning with noisy labels.

\noindent\textbf{Explicit Extraction Branch.} 
While the implicit branch captures structural semantics indirectly, its performance may deteriorate under domain shifts due to limited sensitivity to distributional changes. To enhance structural awareness, we introduce a complementary branch that explicitly encodes topological information by extracting high-order subgraph patterns~\cite{shervashidze2011weisfeiler,nikolentzos2021graph}. This design enables the model to generate graph-level representations that are more robust to structural discrepancies across domains. Specifically, we formulate the explicit extraction branch as:
\begin{equation}
\begin{aligned}
\mathbf{h}_v &= \phi\left( \mathcal{S}_v\left(G\right) \right), \quad \forall v \in \mathcal{V},
\\
\mathbf{z}_G^{EB} &= \operatorname{READOUT}\left( \left\{ \mathbf{h}_v\right\}_{v\in \mathcal{V}} \right),
\nonumber
\end{aligned}
\end{equation}
where $\mathcal{S}_v(G)$ denotes a set of high-order substructures extracted from $G$ (e.g., shortest paths~\cite{borgwardt2005shortest} or subtree patterns~\cite{shervashidze2011weisfeiler}), $\phi(\cdot)$ encodes each substructure into a latent representation, and $\operatorname{READOUT}(\cdot)$ aggregates these representations into a graph-level embedding. The resulting $\mathbf{z}_G^{EB}$ serves as the explicit topological representation of the graph.

\noindent\textbf{Noise-Resilient Pre-Training.}
To mitigate the impact of label noise in the source domain, we exploit local semantic consistency among graphs in the feature space. Empirically, semantically similar graphs tend to exhibit stable feature distributions, even under corrupted labels~\cite{wang2020understanding,iscen2022learning}. Based on this insight, we construct a semantic neighbor graph by identifying the top-$k$ nearest neighbors for each source sample using similarity $\alpha_{ij}={{\mathbf{z}_{G_i}^B}^\top\mathbf{z}_{G_j}^B}/{||\mathbf{z}_{G_i}^B||\cdot||\mathbf{z}_{G_j}^B||}$ over graph-level embeddings obtained from each branch, where $B\in \{\text{IB}, \text{EB}\}$.
To enforce prediction consistency within local neighborhoods, we encourage the predicted distribution to align with a weighted average of its semantic neighbors’ predictions: 
\begin{equation}
    \mathcal{L}_{\text{noise}}^B = \frac{1}{n_s} \sum_{i=1}^{n_s} {\text{KL}}\left( {\mathbf{z}_{G_i}^B}\Bigg\| \sum_{j \in top-k(G_i)} \alpha_{ij}\cdot \mathbf{z}_{G_j}^B \right),\nonumber
\end{equation}
where ${\text{KL}}$ is the Kullback-Leibler divergence, $top-k(G_i)$ is the top-$k$ nearest neighbors samples of $G_i$. This regularization guides the model to learn noise-resilient representations by aligning each prediction with its semantic context, rather than relying solely on potentially corrupted labels. In formulation, we pre-train the dual branches with:
\begin{equation}
\label{pre-train}
    \mathcal{L}_{\text{pre}}^B=\mathcal{L}_{\text{sup}}^B+\beta\mathcal{L}_{\text{noise}}^B,
\end{equation}
where $\mathcal{L}_{\text{sup}}^B=\frac{1}{n_s}\sum_{i=1}^{n_s}l(\sigma(\mathbf{z}_{G_i}^B),y_i)$ is the supervised classification loss, $l$ is the cross-entropy loss and $\sigma$ is the softmax function. $B\in\{\text{IB}, \text{EB}\}$ indicates the implicit and explicit branches pre-training.

\subsection{Nested Pseudo-Label Refinement}
While various domain adaptation techniques such as distribution alignment~\cite{long2015learning,ganin2016domain} and adversarial training~\cite{tzeng2017adversarial,pei2018multi} have been widely explored, they often rely on strong assumptions regarding the existence of domain-invariant representations, which may not hold in the presence of label noise. In contrast, pseudo-labeling provides a flexible and data-driven alternative by leveraging model predictions on unlabeled target samples to guide adaptation~\cite{lee2013pseudo,chen2020selftraining}. In our setting, the dual-branch encoder offers two complementary perspectives for estimating target semantics, enabling more reliable pseudo-label selection through confidence-based filtering. This design facilitates progressive adaptation by gradually incorporating trustworthy target samples into training, while retaining the robustness of the noise-resilient pre-trained branches.

Specifically, at each iteration of cross-branch pseudo-label refinement, we select one branch to generate predictions for all target domain samples. For each sample $G_j^t \in \mathcal{D}^t$, we compute the predicted class probability vector $\hat{\mathbf{y}}_j = \operatorname{Softmax}(\mathbf{z}_{G_j}^B)$, where $B\in\{\text{IB}, \text{EB}\}$. We then select a set of high-confidence samples $\mathcal{T}_{\text{conf}}$ defined as:
\begin{equation}
\label{filter}
    \mathcal{T}_{\text{conf}}^B = \left\{ G_j^t \in \mathcal{D}^t \mid \max(\hat{\mathbf{y}}_j) \geq \zeta \right\},
\end{equation}
where $\zeta$ is a pre-defined threshold. The corresponding pseudo-labels are assigned as:
$\tilde{y}_j = \arg\max(\hat{\mathbf{y}}_j), \forall G_j^t \in \mathcal{T}_{\text{conf}}^B$.
The selected pseudo-labeled samples $\{(G_j^t, \tilde{y}_j)\}_{j}$ are then used to fine-tune the other branch with:
\begin{equation}
    \mathcal{L}_{\text{refine}}^{B'} = \mathcal{L}_{\text{pre}}^{B'}- \frac{1}{|\mathcal{T}_{\text{conf}}^B|} \sum_{\scriptscriptstyle{G_j^t \in \mathcal{T}_{\text{conf}}^B}} \tilde{y}_j\log \sigma(\mathbf{z}_{G_j^t}^{B'}),
\end{equation}
where $\sigma(\mathbf{z}_{G_j^t}^{B'})$ denotes the predicted probability from the other branch $B'$ and $\sigma$ is the $\operatorname{Softmax}$ operation.
The two branches are alternated in subsequent iterations, allowing the model to progressively adapt through mutual supervision.

\subsection{Noisy Pseudo-Label Tolerated Regularization}
Pseudo-labeling facilitates adaptation to the target domain by providing surrogate supervision, yet it inevitably introduces label noise that may compromise model performance~\cite{rizve2021in}. To address this issue, we propose a noise-aware regularization term that penalizes overconfident or unstable predictions during refinement. This regularization serves as a soft constraint to suppress the influence of unreliable pseudo-labels, guiding the model toward more consistent and robust predictions. Moreover, we provide a theoretical analysis, which guarantees its ability to mitigate the negative impact of noisy supervision and enhance generalization in the target domain. Specifically, we define the refinement loss with the noisy tolerated regularization as:
\begin{equation}
\label{re-loss}
    \mathcal{L}_{\text{Re}}^{B'}\!=\!\mathcal{L}_{\text{refine}}^{B'}-\frac{\lambda}{|\mathcal{T}_{\text{conf}}^B|}\sum_{\scriptscriptstyle{G_j^t \in \mathcal{T}_{\text{conf}}^B}}\log\left(\langle \sigma(\mathbf{z}_{G_j^t}^{B'}),\sigma(\mathbf{z}_{G_j^t}^B)\rangle\right),
\end{equation}
where $\langle \sigma(\mathbf{z}_{G_j^t}^{B'}), \sigma(\mathbf{z}_{G_j^t}^{B}) \rangle$ denotes the inner product between the softmax predictions of the two branches. 
Here, $\sigma$ represents the $\operatorname{Softmax}$ function, and $\mathbf{z}_{G_j^t}^{B'}$, $\mathbf{z}_{G_j^t}^{B}$ are the graph-level embeddings of $G_j^t$ produced by branches $B'$ and $B$, respectively. $B, B' \in \{\text{IB}, \text{EB}\}$ with $B \neq B'$. For future analysis of the effectiveness of noisy-tolerant regularization, we derive the gradient of Eq.~\eqref{re-loss} and introduce Lemma~\ref{lemma1}.
\begin{lemma}
\label{lemma1}
Let $\Theta$ denote the parameters of branch $B'$. The gradient of Eq.~\ref{re-loss} with respect to $\Theta$ is given by:
\begin{equation}
    \nabla_\Theta \mathcal{L}_{\text{Re}}^{B'} = \frac{1}{|\mathcal{T}_{\text{conf}}^B|} \sum_{\scriptscriptstyle G_j^t \in \mathcal{T}_{\text{conf}}^B} 
\nabla_\Theta \mathbf{z}_{G_j^t}^{B'} \cdot 
\left( \mathbf{p}_j - \tilde{{y}}_j + \lambda \cdot \mathbf{g}_j \right),\nonumber
\end{equation}
where $\mathbf{p}_j=\sigma(\mathbf{z}_{G_j^t}^{B'})$, $\mathbf{q}_j=\sigma(\mathbf{z}_{G_j^t}^B)$, and the regularization gradient $\mathbf{g}_{j} \in \mathbb{R}^C$ is defined as:
\begin{equation}
\begin{aligned}
    \mathbf{g}_j :&= \frac{1}{\langle \mathbf{p}_j, \mathbf{q}_j \rangle} \cdot \mathbf{J}_{\mathbf{p}_j}^\top \mathbf{q}_j,\\
    \text{with}\quad
    [\mathbf{J}_{\mathbf{p}_j}]_{ck} &= \frac{\partial p_{j,c}}{\partial z_{j,k}^{B'}} = p_{j,c} (\delta_{ck} - p_{j,k}).\nonumber
\end{aligned}
\end{equation}
Here, $\delta_{ck}$ denotes the Kronecker delta, which equals 1 if $c = k$ and 0 otherwise.
\end{lemma}

\begin{algorithm}[!htb]
    \caption{Nested Pseudo-Label Refinement (\method{})}
    
    \renewcommand{\algorithmicrequire}{\textbf{Input:}} 
    \renewcommand{\algorithmicensure}{\textbf{Output:}} 
    \begin{algorithmic}[1]
        \REQUIRE 
        Source domain data $\mathcal{D}_s = \{(G_i^s, y_i^s)\}$, target domain data $\mathcal{D}_t = \{G_j^t\}$, number of iterations $T$
        \ENSURE 
        Trained model parameters $\Theta$ for implicit branch (\text{IB}) and $\Theta'$ for explicit branch (\text{EB})
        
\texttt{/Stage 1: Dual Branches Pre-Training/}
        \FOR{$B,B'\in\{\text{IB},\text{EB}\}, B\neq B'$}
            \STATE Update $\Theta$ with Eq.~\eqref{pre-train}
            \STATE Update $\Theta'$ with Eq.~\eqref{pre-train}
        \ENDFOR
        
        \texttt{/Stage 2: Nested Refinement with Regularization/} 
        \FOR{$i=1$ to $T$}
            \STATE Filter high-confidence samples $\mathcal{T}_{\text{conf}}^B$ from branch $B$ with Eq.\eqref{filter}
            \STATE Update $\Theta'$ of $\text{EB}$ branch by Eq.~\eqref{re-loss}
            \STATE Filter high-confidence samples $\mathcal{T}_{\text{conf}}^{B'}$ from branch $B'$ with Eq.\eqref{filter}
            \STATE Update $\Theta$ of $\text{IB}$ branch by Eq.~\eqref{re-loss}
        \ENDFOR
       \STATE \textbf{return} Dual branches parameters $\Theta$ and $\Theta'$ 
    \end{algorithmic}
    \label{algorithm1}
    \vspace{-2pt}
\end{algorithm}

From Lemma~\ref{lemma1}, we observe that when the pseudo label $\tilde{y}_j$ is correct, the prediction $\mathbf{p}_j$ increasingly aligns with it during training, causing the cross-entropy gradient to diminish. This reduction weakens the learning signal from clean samples and allows noisy examples to dominate the optimization. The regularization term $\mathbf{g}_j$ alleviates this issue by maintaining substantial gradient contributions for clean instances, thus preserving their supervisory effect even as the loss converges.
When $\tilde{y}_j$ is incorrect, the cross-entropy term $\mathbf{p}_{j} - \tilde{y}_{j}$ becomes positive, leading to updates that push the model away from the true class. The regularization term $\mathbf{g}_{j}$, which is typically negative at the true class index, counteracts this effect by reducing the gradient magnitude on mislabeled examples. This dampening mechanism limits the influence of noisy labels during optimization.

\subsection{Learning Framework}
The overall learning framework is outlined in Algorithm~\ref{algorithm1}, which adopts an alternating dual-branch strategy to progressively refine pseudo labels and suppress the influence of label noise. The process begins with noise-resilient pre-training on the source domain to initialize both the implicit and explicit branches (lines 1–3). At each iteration, one branch generates pseudo labels for the target domain, and high-confidence samples are selected based on prediction probability (lines 6 and 8). These samples are then used to update the other branch via a regularized training objective (lines 7 and 9). The two branches alternate roles throughout training (lines 5–10), enabling mutual correction and promoting robust adaptation under noisy supervision.




\begin{table*}[t]
\small
        \caption{The graph 
 classification results (in \%) on the PROTEINS dataset under graph flux density domain shift (source $\rightarrow$ target).  P0, P1, P2 and P3 denote the sub-datasets partitioned with graph flux density. \textbf{Bold} results indicate the best performance.}  
        \label{tab:proteins_flux}
	\resizebox{1.0\textwidth}{!}{
	\begin{tabular}{l|c|c|c|c|c|c|c|c|c|c|c|c}
		\toprule                                                      
		\textbf{Methods}    & P0$\rightarrow$P1 & P1$\rightarrow$P0 & P0$\rightarrow$P2 & P2$\rightarrow$P0 & P0$\rightarrow$P3 & P3$\rightarrow$P0 & P1$\rightarrow$P2 & P2$\rightarrow$P1 & P1$\rightarrow$P3 & P3$\rightarrow$P1 & P2$\rightarrow$P3 & P3$\rightarrow$P2
  \\ \midrule
		WL  & 67.5\scriptsize{$\pm$1.4} & 31.9\scriptsize{$\pm$1.9} & 54.7\scriptsize{$\pm$0.8} & 67.0\scriptsize{$\pm$1.5} & 24.2\scriptsize{$\pm$2.4} & 21.6\scriptsize{$\pm$1.8} & 49.8\scriptsize{$\pm$1.0} & 43.3\scriptsize{$\pm$1.7} & 33.4\scriptsize{$\pm$1.9} & 61.2\scriptsize{$\pm$1.3} & 32.9\scriptsize{$\pm$0.8} & 43.6\scriptsize{$\pm$2.1} \\
        PathNN      & 68.0\scriptsize{$\pm$1.4} & 72.6\scriptsize{$\pm$2.6} & 55.1\scriptsize{$\pm$2.3} & 38.2\scriptsize{$\pm$2.8} & 25.4\scriptsize{$\pm$2.5} & 22.6\scriptsize{$\pm$4.6} & 39.9\scriptsize{$\pm$3.1} & 63.6\scriptsize{$\pm$1.7} & 34.4\scriptsize{$\pm$2.5} & 27.6\scriptsize{$\pm$2.2} & 67.0\scriptsize{$\pm$1.9} & 46.7\scriptsize{$\pm$2.0} \\
        GCN         & 67.3\scriptsize{$\pm$3.5} & 73.3\scriptsize{$\pm$4.3} & 55.9\scriptsize{$\pm$1.7} & 72.1\scriptsize{$\pm$2.6} & 23.8\scriptsize{$\pm$1.7} & 22.5\scriptsize{$\pm$1.4} & 52.3\scriptsize{$\pm$3.9} & 63.9\scriptsize{$\pm$2.4} & 27.3\scriptsize{$\pm$1.0} & 45.6\scriptsize{$\pm$1.7} & 30.3\scriptsize{$\pm$2.1} & 47.7\scriptsize{$\pm$1.4}  \\
        GIN         & 62.3\scriptsize{$\pm$2.3} & 59.5\scriptsize{$\pm$2.5} & 50.6\scriptsize{$\pm$2.1} & 49.4\scriptsize{$\pm$2.4} & 24.8\scriptsize{$\pm$1.3} & 60.0\scriptsize{$\pm$0.9} & 45.2\scriptsize{$\pm$0.3} & 56.4\scriptsize{$\pm$3.1} & 66.0\scriptsize{$\pm$1.2} & 34.3\scriptsize{$\pm$1.7} & 33.4\scriptsize{$\pm$1.4} & 48.5\scriptsize{$\pm$1.9}  \\
        GAT         & 62.8\scriptsize{$\pm$0.8} & 68.1\scriptsize{$\pm$1.2} & 50.1\scriptsize{$\pm$1.7} & 66.2\scriptsize{$\pm$1.4} & 64.6\scriptsize{$\pm$2.3} & 18.0\scriptsize{$\pm$1.4} & 48.9\scriptsize{$\pm$1.0} & 62.8\scriptsize{$\pm$1.8} & 46.5\scriptsize{$\pm$1.4} & 25.5\scriptsize{$\pm$1.1} & 33.1\scriptsize{$\pm$0.9} & 49.0\scriptsize{$\pm$2.7}  \\
        GMT         & 49.6\scriptsize{$\pm$1.0} & 51.3\scriptsize{$\pm$1.3} & 54.1\scriptsize{$\pm$1.6} & 50.6\scriptsize{$\pm$1.3} & 53.8\scriptsize{$\pm$1.1} & 51.4\scriptsize{$\pm$1.7} & 52.9\scriptsize{$\pm$1.9} & 53.0\scriptsize{$\pm$1.1} & 53.5\scriptsize{$\pm$1.0} & 50.4\scriptsize{$\pm$1.1} & 52.5\scriptsize{$\pm$1.2} & 50.2\scriptsize{$\pm$1.0} \\
        \midrule
        Co-teaching & 67.4\scriptsize{$\pm$0.5} & 69.2\scriptsize{$\pm$1.2} & 54.2\scriptsize{$\pm$1.7} & 69.4\scriptsize{$\pm$0.4} & 24.7\scriptsize{$\pm$1.9} & 25.5\scriptsize{$\pm$1.3} & 49.4\scriptsize{$\pm$0.8} & 61.4\scriptsize{$\pm$2.6} & 38.9\scriptsize{$\pm$2.1} & 47.4\scriptsize{$\pm$2.5} & 43.0\scriptsize{$\pm$1.8} & 46.4\scriptsize{$\pm$3.3} \\
        Taylor-CE   & 65.7\scriptsize{$\pm$3.6} & 66.4\scriptsize{$\pm$4.3} & 49.3\scriptsize{$\pm$3.5} & 53.6\scriptsize{$\pm$2.9} & 27.9\scriptsize{$\pm$1.5} & 57.4\scriptsize{$\pm$2.4} & 50.6\scriptsize{$\pm$2.2} & 42.7\scriptsize{$\pm$1.8} & 69.7\scriptsize{$\pm$1.9} & 39.6\scriptsize{$\pm$1.7} & 40.4\scriptsize{$\pm$1.3} & 42.0\scriptsize{$\pm$2.7} \\
        RTGNN       & 63.0\scriptsize{$\pm$1.8} & 70.3\scriptsize{$\pm$1.2} & 61.1\scriptsize{$\pm$1.8} & 67.7\scriptsize{$\pm$2.5} & 26.0\scriptsize{$\pm$0.7} & 20.0\scriptsize{$\pm$0.9} & 55.1\scriptsize{$\pm$1.4} & 67.3\scriptsize{$\pm$1.7} & 24.4\scriptsize{$\pm$1.3} & 48.9\scriptsize{$\pm$1.5} & 34.8\scriptsize{$\pm$1.2} & 44.0\scriptsize{$\pm$1.5} \\
        OMG         & 64.9\scriptsize{$\pm$1.4} & 72.2\scriptsize{$\pm$1.7} & 47.1\scriptsize{$\pm$1.1} & 63.3\scriptsize{$\pm$1.9} & 68.1\scriptsize{$\pm$1.3} & 22.3\scriptsize{$\pm$0.8} & 46.3\scriptsize{$\pm$2.3} & 59.3\scriptsize{$\pm$2.2} & 52.5\scriptsize{$\pm$1.8} & 21.8\scriptsize{$\pm$1.9} & 35.1\scriptsize{$\pm$1.5} & 43.6\scriptsize{$\pm$1.3} \\
        SPORT       & 60.7\scriptsize{$\pm$1.4} & 65.4\scriptsize{$\pm$1.8} & 49.0\scriptsize{$\pm$1.2} & 69.1\scriptsize{$\pm$0.5} & 54.7\scriptsize{$\pm$1.1} & 51.8\scriptsize{$\pm$1.5} & 55.3\scriptsize{$\pm$2.1} & 64.3\scriptsize{$\pm$2.4} & 51.6\scriptsize{$\pm$1.3} & 25.8\scriptsize{$\pm$1.2} & 34.1\scriptsize{$\pm$1.7} & 42.3\scriptsize{$\pm$1.9} \\ \midrule
        CoCo        & 66.9\scriptsize{$\pm$1.3} & 50.9\scriptsize{$\pm$1.9} & 55.2\scriptsize{$\pm$1.5} & 64.4\scriptsize{$\pm$1.4} & 71.4\scriptsize{$\pm$1.7} & 25.9\scriptsize{$\pm$1.2} & 51.6\scriptsize{$\pm$2.6} & 55.1\scriptsize{$\pm$2.4} & 36.7\scriptsize{$\pm$1.8} & 56.3\scriptsize{$\pm$1.2} & 38.3\scriptsize{$\pm$1.9} & 44.5\scriptsize{$\pm$3.0} \\
        DEAL       & 66.7\scriptsize{$\pm$}2.3 & 71.6\scriptsize{$\pm$}2.1 & 55.2\scriptsize{$\pm$}1.9 & 70.4\scriptsize{$\pm$}3.0 & 34.7\scriptsize{$\pm$}1.0 & 58.6\scriptsize{$\pm$}1.7 & 51.0\scriptsize{$\pm$}2.0 & 65.3\scriptsize{$\pm$}1.6 & 43.7\scriptsize{$\pm$}1.8 & 66.5\scriptsize{$\pm$}1.9 & 63.4\scriptsize{$\pm$}3.1 & 46.4\scriptsize{$\pm$}2.3 \\
        SGDA        & 67.8\scriptsize{$\pm$}2.1 & 59.4\scriptsize{$\pm$}1.3 & 57.7\scriptsize{$\pm$}1.6 & 73.1\scriptsize{$\pm$}1.8 & 38.3\scriptsize{$\pm$}2.4 & 31.9\scriptsize{$\pm$}2.7 & 48.2\scriptsize{$\pm$}2.0 & 48.8\scriptsize{$\pm$}2.2 & 39.2\scriptsize{$\pm$}2.0 & 58.6\scriptsize{$\pm$}1.6 & 40.2\scriptsize{$\pm$}1.8 & 46.8\scriptsize{$\pm$}2.3 \\
        A2GNN       & 60.7\scriptsize{$\pm$}2.2 & 65.5\scriptsize{$\pm$}1.8 & 54.3\scriptsize{$\pm$}2.0 & 67.5\scriptsize{$\pm$}2.2 & 60.2\scriptsize{$\pm$}1.9 & 53.3\scriptsize{$\pm$}1.7 & 44.2\scriptsize{$\pm$}1.5 & 63.1\scriptsize{$\pm$}1.8 & 42.9\scriptsize{$\pm$}2.3 & 35.7\scriptsize{$\pm$}2.5 & 46.5\scriptsize{$\pm$}2.0 & 53.8\scriptsize{$\pm$}2.1  \\
        StruRW      &  62.5\scriptsize{$\pm$}2.1 & 72.9\scriptsize{$\pm$}1.4 & 59.2\scriptsize{$\pm$}1.8 & 71.0\scriptsize{$\pm$}2.0 & 39.8\scriptsize{$\pm$}1.9 & 34.9\scriptsize{$\pm$}2.1 & 49.6\scriptsize{$\pm$}1.6 & 66.6\scriptsize{$\pm$}2.1 & 37.4\scriptsize{$\pm$}2.3 & 61.1\scriptsize{$\pm$}1.7 & 40.5\scriptsize{$\pm$}1.5 & 45.9\scriptsize{$\pm$}2.2 \\
        PA-BOTH     & 64.9\scriptsize{$\pm$}1.7 & 73.6\scriptsize{$\pm$}2.1 & 58.0\scriptsize{$\pm$}2.2 & 69.1\scriptsize{$\pm$}1.9 & 36.5\scriptsize{$\pm$}2.3 & 54.3\scriptsize{$\pm$}1.5 & 53.9\scriptsize{$\pm$}1.8 & 67.2\scriptsize{$\pm$}1.4 & 42.2\scriptsize{$\pm$}1.6 & 67.6\scriptsize{$\pm$}2.0 & 63.1\scriptsize{$\pm$}1.9 & 45.3\scriptsize{$\pm$}2.1  \\ \midrule
        ROAD & 52.2\scriptsize{$\pm$2.6} & 53.8\scriptsize{$\pm$3.2} & 60.9\scriptsize{$\pm$2.7} & 55.9\scriptsize{$\pm$2.1} & 63.1\scriptsize{$\pm$2.0} & 57.2\scriptsize{$\pm$2.7} & 58.6\scriptsize{$\pm$2.4} & 58.2\scriptsize{$\pm$1.7} & 62.5\scriptsize{$\pm$2.0} & 58.2\scriptsize{$\pm$1.8} & 61.1\scriptsize{$\pm$2.5} & 57.2\scriptsize{$\pm$1.7} \\
ALEX & 68.7\scriptsize{$\pm$2.7} & \textbf{74.9\scriptsize{$\pm$3.0}} & 62.5\scriptsize{$\pm$2.8} & 68.6\scriptsize{$\pm$2.6} & 73.7\scriptsize{$\pm$2.8} & 61.3\scriptsize{$\pm$3.4} & 62.8\scriptsize{$\pm$2.6} & 64.9\scriptsize{$\pm$2.1} & 68.2\scriptsize{$\pm$2.0} & 61.7\scriptsize{$\pm$2.2} & 64.1\scriptsize{$\pm$3.0} & 58.0\scriptsize{$\pm$2.2} \\
\midrule
        \method{}        & \textbf{71.7\scriptsize{$\pm$2.4}} & 74.7\scriptsize{$\pm$2.6} & \textbf{64.5\scriptsize{$\pm$2.1}} & \textbf{73.3\scriptsize{$\pm$2.1}} & \textbf{77.1\scriptsize{$\pm$2.4}} & \textbf{63.2\scriptsize{$\pm$1.7}} & \textbf{63.8\scriptsize{$\pm$2.5}} & \textbf{68.1\scriptsize{$\pm$2.2}} & \textbf{70.5\scriptsize{$\pm$2.1}} & \textbf{68.4\scriptsize{$\pm$2.4}} & \textbf{67.2\scriptsize{$\pm$2.3}} & \textbf{61.0\scriptsize{$\pm$1.6}}       \\ \bottomrule
	\end{tabular}}
\end{table*}

\begin{table}[t]
\centering
\small
\caption{The graph classification results (in \%) under semantic information shift (source$\rightarrow$target). P, D, C, CM, B, and BM denote PROTEINS, DD, COX2, COX2\_MD, BZR, and BZR\_MD, respectively. \textbf{Bold} indicates the best performance. OOM means out of memory.}
\resizebox{0.475\textwidth}{!}{
\begin{tabular}{l|c|c|c|c|c|c}
\toprule
\textbf{Methods} & P$\rightarrow$D & D$\rightarrow$P & C$\rightarrow$CM & CM$\rightarrow$C & B$\rightarrow$BM & BM$\rightarrow$B \\
\midrule
WL& 42.5\scriptsize{$\pm$2.0} & 43.6\scriptsize{$\pm$2.4} & 50.7\scriptsize{$\pm$1.5} & 54.8\scriptsize{$\pm$2.0} & 50.6\scriptsize{$\pm$2.2} & 25.3\scriptsize{$\pm$2.3} \\
PathNN & 47.5\scriptsize{$\pm$1.5} & 41.1\scriptsize{$\pm$2.0} & 49.8\scriptsize{$\pm$1.6} & 66.9\scriptsize{$\pm$2.6} & 50.3\scriptsize{$\pm$1.6} & 37.2\scriptsize{$\pm$2.4} \\
GCN & 53.7\scriptsize{$\pm$2.3} & 51.8\scriptsize{$\pm$2.0} & 49.8\scriptsize{$\pm$1.6} & 32.7\scriptsize{$\pm$2.9} & 49.7\scriptsize{$\pm$2.1} & 55.5\scriptsize{$\pm$2.7} \\
GIN & 48.3\scriptsize{$\pm$1.9} & 49.9\scriptsize{$\pm$1.7} & 51.2\scriptsize{$\pm$2.0} & 52.6\scriptsize{$\pm$2.5} & 48.7\scriptsize{$\pm$2.0} & 55.8\scriptsize{$\pm$1.9} \\
GAT & 59.2\scriptsize{$\pm$1.7} & 57.4\scriptsize{$\pm$2.0} & 49.3\scriptsize{$\pm$2.1} & 36.4\scriptsize{$\pm$2.5} & 51.3\scriptsize{$\pm$1.9} & 32.7\scriptsize{$\pm$2.0} \\
GMT & 55.7\scriptsize{$\pm$2.5} & 53.9\scriptsize{$\pm$2.6} & 50.7\scriptsize{$\pm$2.1} & 44.4\scriptsize{$\pm$1.9} & 49.2\scriptsize{$\pm$1.7} & 32.7\scriptsize{$\pm$2.2} \\
\midrule
Co-teaching & 55.9\scriptsize{$\pm$2.2} & 60.1\scriptsize{$\pm$1.8} & 47.7\scriptsize{$\pm$2.3} & 48.8\scriptsize{$\pm$2.0} & 50.8\scriptsize{$\pm$2.4} & 44.2\scriptsize{$\pm$1.9} \\
Taylor-CE & 55.2\scriptsize{$\pm$2.0} & 55.7\scriptsize{$\pm$2.2} & 51.2\scriptsize{$\pm$1.8} & 55.6\scriptsize{$\pm$2.5} & 48.7\scriptsize{$\pm$2.0} & 44.2\scriptsize{$\pm$1.9} \\
RTGNN & 53.7\scriptsize{$\pm$2.0} & 52.6\scriptsize{$\pm$1.9} & 51.2\scriptsize{$\pm$2.0} & 54.3\scriptsize{$\pm$1.6} & 49.2\scriptsize{$\pm$2.8} & 55.5\scriptsize{$\pm$2.3} \\
OMG & 56.7\scriptsize{$\pm$1.7} & 53.4\scriptsize{$\pm$2.2} & 54.5\scriptsize{$\pm$1.8} & 57.3\scriptsize{$\pm$2.7} & 50.8\scriptsize{$\pm$2.0} & 59.3\scriptsize{$\pm$2.3} \\
SPORT & OOM & OOM & 53.7\scriptsize{$\pm$2.1} & 63.9\scriptsize{$\pm$3.3} & 51.4\scriptsize{$\pm$2.6} & 65.8\scriptsize{$\pm$3.0} \\
\midrule
CoCo & 62.6\scriptsize{$\pm$2.5} & 67.1\scriptsize{$\pm$2.0} & 56.8\scriptsize{$\pm$2.5} & 67.0\scriptsize{$\pm$2.8} & 50.5\scriptsize{$\pm$2.0} & 79.3\scriptsize{$\pm$2.2} \\ 
DEAL & 69.7\scriptsize{$\pm$1.9} & 60.0\scriptsize{$\pm$2.5} & 52.7\scriptsize{$\pm$2.1} & 60.4\scriptsize{$\pm$2.2} & 52.4\scriptsize{$\pm$2.9} & 68.6\scriptsize{$\pm$2.8} \\
SGDA & 53.3\scriptsize{$\pm$1.9} & 55.2\scriptsize{$\pm$3.3} & 54.1\scriptsize{$\pm$2.8} & 52.6\scriptsize{$\pm$2.7} & 49.6\scriptsize{$\pm$2.4} & 48.3\scriptsize{$\pm$2.1} \\
A2GNN & 61.6\scriptsize{$\pm$2.9} & 68.8\scriptsize{$\pm$2.7} & 51.2\scriptsize{$\pm$2.0} & 65.4\scriptsize{$\pm$2.5} & 52.1\scriptsize{$\pm$2.7} & 61.1\scriptsize{$\pm$2.7} \\
StruRW & 52.8\scriptsize{$\pm$1.9} & 56.4\scriptsize{$\pm$3.3} & 52.8\scriptsize{$\pm$2.8} & 51.3\scriptsize{$\pm$2.7} & 48.7\scriptsize{$\pm$2.4} & 49.7\scriptsize{$\pm$3.1} \\
PA-BOTH & 56.5\scriptsize{$\pm$2.9} & 54.2\scriptsize{$\pm$2.6} & 51.2\scriptsize{$\pm$2.9} & 58.9\scriptsize{$\pm$2.3} & 48.7\scriptsize{$\pm$2.7} & 47.7\scriptsize{$\pm$2.5} \\
\midrule
ROAD & 55.2\scriptsize{$\pm$2.0} & 59.5\scriptsize{$\pm$3.0} & 55.2\scriptsize{$\pm$2.6} & 70.2\scriptsize{$\pm$2.7} & 52.7\scriptsize{$\pm$2.1} & 79.0\scriptsize{$\pm$2.3} \\
ALEX & 68.8\scriptsize{$\pm$2.1} & 68.1\scriptsize{$\pm$2.2} & 56.2\scriptsize{$\pm$2.0} & 69.2\scriptsize{$\pm$2.9} & 54.3\scriptsize{$\pm$2.1} & 78.8\scriptsize{$\pm$3.0} \\
\midrule
\method{} & \textbf{72.3\scriptsize{$\pm$2.6}} & \textbf{69.9\scriptsize{$\pm$2.8}} & \textbf{57.3\scriptsize{$\pm$2.6}} & \textbf{73.0\scriptsize{$\pm$2.3}} & \textbf{55.9\scriptsize{$\pm$3.0}} & \textbf{80.0\scriptsize{$\pm$2.7}} \\
\bottomrule
\end{tabular}
}
\label{tab:semantic_shift}
\end{table}

\section{Experiments}

\subsection{Experimental Settings}
\textbf{Datasets.} To assess the effectiveness of the proposed \method{}, we conduct extensive experiments on multiple benchmark datasets from TUDataset, covering diverse types of domain shifts. For structure-based domain shifts, we utilize MUTAGENICITY~\cite{kazius2005derivation}, NCI1~\cite{wale2008comparison}, FRANKENSTEIN~\cite{orsini2015graph}, and PROTEINS~\cite{dobson2003distinguishing}, where each dataset is partitioned into source and target domains based on variations in edge, node and graph flux density to simulate structural distribution shifts~\cite{yin2023coco}. For feature-based domain shifts, we evaluate \method{} on PROTEINS, DD, BZR, BZR\_MD, COX2, and COX2\_MD~\cite{sutherland2003spline}, where domain discrepancies primarily arise from differences in semantic feature distributions. Detailed dataset statistics are provided in Appendix C.

\noindent\textbf{Baselines.} We compare the proposed \method{} with a comprehensive set of competitive baselines on the datasets above. These baselines include two graph kernel methods: WL~\cite{shervashidze2011weisfeiler} and PathNN~\cite{michel2023path}; four general graph neural networks: GCN~\cite{kipf2022semi}, GIN~\cite{xu2018powerful}, GAT~\cite{velivckovic2018graph}, and GMT~\cite{baek2021accurate}; five label denoising methods: Co-teaching~\cite{han2018co}, RTGNN~\cite{qian2023robust}, Taylor-CE~\cite{feng2021can}, OMG~\cite{yin2023omg}, and SPORT~\cite{yin2024sport}; six graph domain adaptation methods: DEAL~\cite{yin2022deal}, CoCo~\cite{yin2023coco}, SGDA~\cite{qiao2023semi}, A2GNN~\cite{liu2024rethinking}, StruRW~\cite{liu2023structural}, and PA-BOTH~\cite{liu2024pairwise}; and two methods that address both label noise and domain adaptation: ALEX~\cite{yuan2023alex} and ROAD~\cite{feng2023road}. More detailed descriptions of the baseline settings are provided in Appendix D.

\noindent\textbf{Implementation Details.} We implement \method{} and all baseline models using PyTorch and conduct all experiments on NVIDIA A100 GPUs to ensure a fair comparison. For \method{}, the implicit branch (IB) is instantiated with the GMT~\cite{baek2021accurate} architecture to capture semantic consistency via message passing, while the explicit branch (EB) employs the PathNN model~\cite{michel2023path} to extract high-order topological structures explicitly. Both branches use 4 GNN layers, with a hidden dimension of 256 and a weight decay of $10^{-12}$. The models are trained using the Adam optimizer with a learning rate of $10^{-4}$. All the models are trained on noisy labeled source graphs and evaluated on unlabeled target graphs. We set the noise ratio $\alpha = 0.3$ and the pseudo-label threshold $\zeta = 0.9$ by default. All reported results are averaged over five independent runs.

\subsection{Performance Comparison}

\begin{table*}[t]
        \caption{The results of ablation studies on the PROTEINS dataset (source → target). \textbf{Bold} results indicate the best performance.}  
	\resizebox{\textwidth}{!}{
	\begin{tabular}{l|c|c|c|c|c|c|c|c|c|c|c|c}
		\toprule                                                      
		\textbf{Methods}    & P0→P1 & P1→P0 & P0→P2 & P2→P0 & P0→P3 & P3→P0 & P1→P2 & P2→P1 & P1→P3 & P3→P1 & P2→P3 & P3→P2  \\ \midrule
		{\method{} w/o IB} & 50.8 & 50.6& 50.7 & 52.2& 50.0 & 48.6& 52.0 & 52.2 & 47.7 & 50.8 & 52.5 & 50.3 \\
		{\method{} w/o EB} & 50.4 & 52.1 & 49.0 & 52.1 & 49.0 & 51.6 & 46.5 & 50.4 & 51.6 & 50.4 & 53.3 & 49.9 \\
            {\method{} w/o NRL} & 68.5 & 71.4 & 62.5 & 70.0 & 73.6 & 60.9 & 61.2 & 65.2 & 68.4 & 64.6 & 63.8 &58.4 \\
		{\method{} w/o NTR} & 69.7 & 71.0 & 63.8 & 70.3 & 74.8 & 62.4 & 62.4 & 66.2 & 69.0 & 65.5 & 65.8 & 58.9 \\
        \midrule
		\method{}  &\textbf{71.7} & \textbf{74.7} & \textbf{64.5} & \textbf{73.3} & \textbf{77.1} & \textbf{63.2} & \textbf{63.8} & \textbf{68.1}& \textbf{70.5} & \textbf{68.4} & \textbf{67.2} & \textbf{61.0} \\ \bottomrule
	\end{tabular}}
    \label{tab:ablation_proteins}
\end{table*}


We present the results of the proposed \method{} with all baseline models under the setting of graph domain adaptation on different datasets in Tables \ref{tab:proteins_flux}, \ref{tab:semantic_shift}, and \ref{tab:proteins_idx}-\ref{tab:mutag_node}. From these tables, we observe that: (1) Label denoising methods consistently outperform general graph-based approaches, as the presence of noisy labels significantly impairs the performance of standard graph models lacking dedicated noise-handling mechanisms. 
(2) Graph domain adaptation methods generally outperform graph-based and label-denoising approaches by effectively mitigating domain distribution shifts. However, their performance may still degrade when source labels are corrupted, highlighting the need for methods that jointly address domain shift and label noise specific for graphs. (3) Label denoising domain adaptation methods demonstrate superior performance over graph domain adaptation methods, which highlights the importance of explicitly addressing label noise alongside domain alignment to enhance model generalization in noisy cross-domain settings. 
(4) The proposed \method{} consistently achieves the highest performance across datasets in most cases, demonstrating its superiority. The outstanding performance is attributed primarily to two factors: (i) the integration of implicit branch and explicit branch enables comprehensive extraction of both structural and semantic features, substantially enhancing representation quality and classification accuracy; and (ii) the nested refinement and noisy tolerated regularization modules jointly promote robust cross-domain adaptation by progressively selecting reliable supervision and suppressing noisy signals. Additional results on other datasets are provided in Appendix E.


\subsection{Ablation Study}\label{sec:ablation}

We conduct ablation studies to examine the contributions of each component in the proposed \method{}: (1) \method{} w/o IB: It removes the implicit extraction branch; (2) \method{} w/o EB: It removes the explicit extraction branch; (3) \method{} w/o NRL: It removes the noise resilient loss in the pretraining stage; (4) \method{} w/o NTR: It remove the noisy pseudo-label tolerated regularization loss during fine-tuning.

Experimental results are reported in Table \ref{tab:ablation_proteins},~\ref{tab:ablation_nci1}-\ref{tab:ablation_mutag}. From the results, we find that: (1) \method{} outperforms \method{} w/o IB and \method{} w/o EB, underscoring the importance of integrating implicit and explicit branches that capture semantic and structural information. Their joint modeling enforces multi-view prediction consistency, providing a robust foundation for effective domain adaptation. (2) \method{} w/o NRL demonstrates inferior performance compared to \method{}. The NRL effectively reduces the negative impact of noisy labels in the source domain by promoting local consistency among neighboring nodes. This constraint enables \method{} to learn noise-resistant representations suitable for domain adaptation. (3) \method{} outperforms \method{} w/o NTR, demonstrating that the noise tolerant regularization effectively mitigates the impact of noisy pseudo-labels by preserving reliable supervision from clean samples. This constraint prevents overfitting and enhances the model’s robustness and generalization across domains. Additional results on other datasets are provided in Appendix E.

\begin{figure}[t]
    \centering
    \hspace{-1.0cm}
    \subfloat[Threshold $\zeta$]{\includegraphics[width=0.265\textwidth,height=0.14\textwidth]{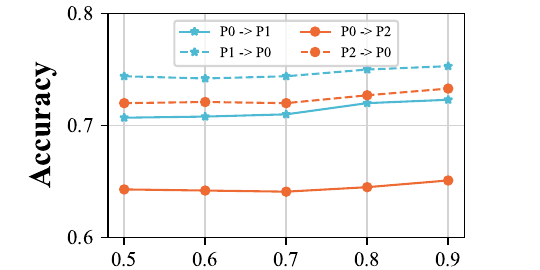}}
    \label{fig:proteins_zeta}
    \hspace{-0.5cm}
    \subfloat[Noise ratio $\alpha$]{\includegraphics[width=0.265\textwidth,height=0.14\textwidth]{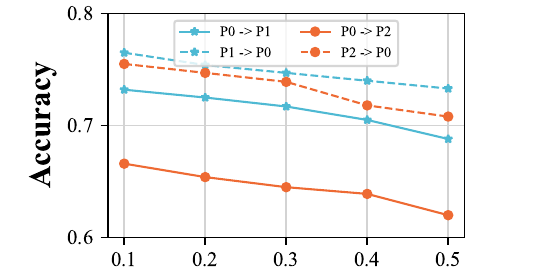}}
    \label{fig:proteins_alpha}
    \hspace{-1.3cm}

    \caption{Hyperparameter sensitivity analysis of threshold $\zeta$ and noise ratio $\alpha$ on the PROTEINS datasets.}
    \label{fig:proteins_hyper}
\end{figure}

\subsection{Sensitivity Analysis}\label{sec:sensitivity}

We perform a sensitivity analysis to examine how the key hyperparameters of \method{}, namely the pseudo-label confidence threshold $\zeta$ and the noise ratio $\alpha$, affect its performance. Specifically, $\zeta$ governs the selection of high-confidence pseudo-labeled target samples, while $\alpha$ determines the proportion of corrupted labels in the source domain. Both parameters play a critical role in balancing supervision quality and model robustness.

Figure \ref{fig:proteins_hyper} illustrates how $\zeta$ and  $\alpha$ affect the performance of \method{} on the PROTEINS dataset.  We vary $\zeta$ within the range of $\{0.5, 0.6, 0.7, 0.8, 0.9\}$ and $\alpha$ in $\{0.1, 0.2, 0.3, 0.4, 0.5\}$. From the results, we observe that: (1) The performance of \method{} in Figure \ref{fig:proteins_hyper}(a) steadily increases as threshold $\zeta$ rises. A higher threshold can effectively filter out pseudo-labels with lower confidence, which reduces the risk of propagating incorrect information during model training, enabling the model to learn from more reliable supervision signals. Thus, we set the threshold $\zeta$ to 0.9 as default to ensure optimal pseudo-label reliability. (2) Figure~\ref{fig:proteins_hyper}(b) illustrates a decreasing accuracy trend with an increasing noise ratio $\alpha$. A higher noise ratio introduces more incorrectly labeled samples into the source domain, thereby degrading the reliability of supervisory signals during training. Consequently, this prevents the model from accurately learning discriminative representations. To maintain a balance between realistic data conditions and robust performance, we set the noise ratio $\alpha$ to 0.3 by default. More results on other datasets are shown in Appendix E.

\begin{figure}[t]
    \centering
    \hspace{-1.0cm}
    \subfloat[\vspace{-2pt}Different backbone for IB]{\includegraphics[width=0.265\textwidth,height=0.14\textwidth]{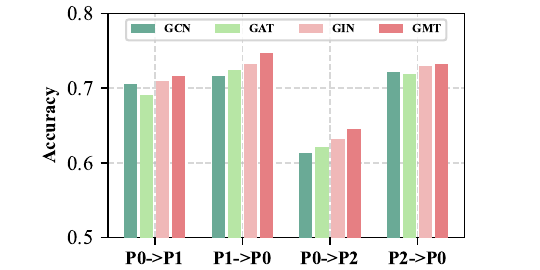}}
    \hspace{-0.5cm}
    \subfloat[Different backbone for EB]{\includegraphics[width=0.265\textwidth,height=0.14\textwidth]{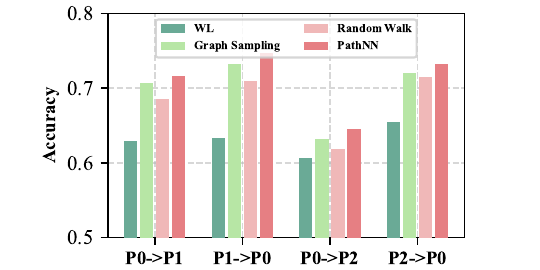}}
    \label{fig:gnn}
    \hspace{-1.3cm}
    \caption{The performance with different backbones for IB and EB on the PROTEINS dataset.}
    \label{fig:gkn}
\end{figure}

\subsection{Flexible Architecture}\label{sec:architecture}
To assess the impact of different backbone choices for the IB and EB branches, we evaluate various message passing methods in IB, including GCN~\cite{kipf2017semi}, GAT~\cite{velickovic2018graph}, GIN~\cite{XuHLJ19}, and GMT~\cite{baek2021accurate}, and adopt several graph kernel-based methods in EB, such as Graph Sampling~\cite{leskovec2006sampling}, Random Walk~\cite{kalofolias2021susan}, WL~\cite{shervashidze2011weisfeiler}, and PathNN~\cite{michel2023path}. As shown in Figure~\ref{fig:gkn}, and consistently observed across other datasets, GMT and PathNN yield the best performance in most cases. This can be attributed to their superior representation capacity, which provides a solid foundation for capturing both semantic and topological features of graphs. These results further validate our choice of GMT in IB and PathNN in EB, as they offer complementary strengths that enhance the effectiveness of dual-branch modeling. Their strong performance also highlights the importance of backbone selection in ensuring stable adaptation under noisy supervision. 
\section{Conclusion}

This paper introduces \method{}, a noise-aware dual-branch framework for robust GDA under label noise. To tackle noisy supervision and distributional shifts, \method{} employs a dual-branch pretraining strategy: one branch captures semantic consistency via local message passing, while the other encodes structural features using a graph kernel method, enabling the extraction of complementary graph representations. A nested pseudo-label refinement mechanism progressively aligns source and target domains by alternately using high-confidence predictions from one branch to supervise the other, enhancing cross-branch consistency and mitigating domain gaps. Additionally, a noise-aware regularization term penalizes overconfident or inconsistent predictions, reducing the impact of noisy labels. Extensive experiments across diverse datasets and noise settings validate the superior robustness and generalization of \method{}, underscoring its promise for reliable graph transfer learning. 

\bibliography{aaai25}

\appendix
\clearpage
\onecolumn
\appendix

\section{A. Proof of Lemma 1}

\noindent \textbf{Lemma 1}
\textit{Let $\Theta$ denote the parameters of branch $B'$. The gradient of the loss function in Eq.~\ref{re-loss} with respect to $\Theta$ is given by
\begin{equation}
    \nabla_\Theta \mathcal{L}_{\text{Re}}^{B'} = \frac{1}{|\mathcal{T}_{\text{conf}}^B|} \sum_{\scriptscriptstyle G_j^t \in \mathcal{T}_{\text{conf}}^B} 
\nabla_\Theta \mathbf{z}_{G_j^t}^{B'} \cdot 
\left( \mathbf{p}_j - \tilde{{y}}_j + \lambda \cdot \mathbf{g}_j \right),\nonumber
\end{equation}
where the regularization gradient $\mathbf{g}_{j} \in \mathbb{R}^C$ is defined as
\begin{equation}
    \mathbf{g}_j := \frac{1}{\langle \mathbf{p}_j, \mathbf{q}_j \rangle} \cdot \mathbf{J}_{\mathbf{p}_j}^\top \mathbf{q}_j,\quad\text{with}\quad
    [\mathbf{J}_{\mathbf{p}_j}]_{ck} = \frac{\partial p_{j,c}}{\partial z_{j,k}^{B'}} = p_{j,c} (\delta_{ck} - p_{j,k}).\nonumber
\end{equation}
Here, $\delta_{ck}$ denotes the Kronecker delta, which equals 1 if $c = k$ and 0 otherwise.}

\textit{Proof:}
\begin{equation}
\begin{aligned}
    \mathcal{L}_{\text{Re}}^{B'}&=\mathcal{L}_{\text{refine}}^{B'}-\frac{\lambda}{|\mathcal{T}_{\text{conf}}^B|}\sum_{\scriptscriptstyle{G_j^t \in \mathcal{T}_{\text{conf}}^B}}\log\left(\langle \sigma(\mathbf{z}_{G_j^t}^{B'}),\sigma(\mathbf{z}_{G_j^t}^B)\rangle\right)\\
      &= \mathcal{L}_{\text{pre}}^{B'}- \frac{1}{|\mathcal{T}_{\text{conf}}^B|} \sum_{\scriptscriptstyle{G_j^t \in \mathcal{T}_{\text{conf}}^B}} \tilde{y}_j\log \sigma(\mathbf{z}_{G_j^t}^{B'})-\frac{\lambda}{|\mathcal{T}_{\text{conf}}^B|}\sum_{\scriptscriptstyle{G_j^t \in \mathcal{T}_{\text{conf}}^B}}\log\left(\langle \sigma(\mathbf{z}_{G_j^t}^{B'}),\sigma(\mathbf{z}_{G_j^t}^B)\rangle\right)\\
      &=\mathcal{L}_{\text{pre}}^{B'}+\mathcal{L}_{\text{rest}}^{B'}.
\end{aligned}
\end{equation}
Due to the pre-training process of dual branches, models are overfitted to the loss $\mathcal{L}_{\text{pre}}^{B'}$, therefore, we simply focus on the remaining term $\mathcal{L}_{\text{rest}}^{B'}$. Denote $\mathbf{p}_j=\sigma(\mathbf{z}_{G_j^t}^{B'})$, $\mathbf{q}_j=\sigma(\mathbf{z}_{G_j^t}^B)$, we have:
\begin{equation}
\nabla_\Theta \mathcal{L}_{\text{rest}}^{B'} = \frac{1}{|\mathcal{T}_{\text{conf}}^B|} \sum_{\scriptscriptstyle{G_j^t \in \mathcal{T}_{\text{conf}}^B}} \nabla_\Theta \mathbf{z}_{G_j^t}^{B'} \cdot \frac{\partial l^{B'}_{G_j^t}}{\partial \mathbf{z}_{G_j^t}^{B'}},
\label{eq:grad_chain}
\end{equation}
where $l^{B'}_{G_j^t}$ is the loss on the sample $G_j^t$ of branch $B'$.
Then, we compute the partial derivative of the loss with respect to $\mathbf{z}_{G_j^t}^{B'}$:
\begin{align}
\frac{\partial l^{B'}_{G_j^t}}{\partial \mathbf{z}_{G_j^t}^{B'}} 
&= \frac{\partial}{\partial \mathbf{z}_{G_j^t}^{B'}} \left( - \tilde{{y}}_j^\top \log \mathbf{p}_j - \lambda \log \langle \mathbf{p}_j, \mathbf{q}_j \rangle \right) \notag \\
&= \mathbf{p}_j - \tilde{{y}}_j + \lambda \cdot \frac{1}{\langle \mathbf{p}_j, \mathbf{q}_j \rangle} \cdot \mathbf{J}_{\mathbf{p}_j}^\top \mathbf{q}_j,
\label{eq:logit_grad}
\end{align}
where $\mathbf{J}_{\mathbf{p}_j}$ is the Jacobian of the softmax function:
\begin{equation}
[\mathbf{J}_{\mathbf{p}_j}]_{ck} = \frac{\partial p_{j,c}}{\partial z_{j,k}^{B'}} = p_{j,c} (\delta_{ck} - p_{j,k}),
\end{equation}
where $\delta_{ck}$ is the Kronecker delta, equal to 1 if $c = k$, and 0 otherwise.
Define the regularizer gradient vector $\mathbf{g}_j \in \mathbb{R}^C$ as:
\begin{equation}
\mathbf{g}_j := \frac{1}{\langle \mathbf{p}_j, \mathbf{q}_j \rangle} \cdot \mathbf{J}_{\mathbf{p}_j}^\top \mathbf{q}_j.
\end{equation}
The $c$-th entry of $\mathbf{g}_j$ can be explicitly expanded as:
\begin{equation}
g_{j,c} = \frac{p_{j,c}}{\langle \mathbf{p}_j, \mathbf{q}_j \rangle} \sum_{k=1}^{C} (q_{j,k} - q_{j,c}) p_{j,k}.
\label{eq:g_comp}
\end{equation}
where $c \in \{1, \dots, C\}$ denotes the class index; $p_{j,c}$ and $q_{j,k}$ are softmax probabilities from branches $B'$ and $B$, respectively.
Substituting Eq.~\eqref{eq:logit_grad} into Eq.~\eqref{eq:grad_chain}, we obtain the final expression:
\begin{equation}
\nabla_\Theta \mathcal{L}_{\text{rest}}^{B'} = \frac{1}{|\mathcal{T}_{\text{conf}}^B|} \sum_{\scriptscriptstyle{G_j^t \in \mathcal{T}_{\text{conf}}^B}} 
\nabla_\Theta \mathbf{z}_{G_j^t}^{B'} \cdot 
\left( \mathbf{p}_j - \tilde{{y}}_j + \lambda \cdot \mathbf{g}_j \right),
\label{eq:final_grad}
\end{equation}
where $\mathbf{g}_j$ is defined as in Eq.~\eqref{eq:g_comp}.


\section{B. Complexity Analysis}

The overall time complexity of the proposed \method{} framework is determined by two main stages: noise-resilient pre-training and nested pseudo-label refinement. In the pre-training phase, the Implicit Branch (IB) incurs a complexity of $ \mathcal{O}(L \cdot (|\mathcal{V}| + |\mathcal{E}|) \cdot d_g) $, while the Explicit Branch (EB) has a complexity of $ \mathcal{O}(|\mathcal{V}| \cdot (|\mathcal{V}| + |\mathcal{E}|)) $ for shortest-path computation, along with an additional $ \mathcal{O}(n_s^2 \cdot d_g) $ for semantic neighbor search, where $ d_g $ is the embedding dimension and $ L $ represents the number of GNN layers. During the refinement stage, the forward pass over $ n_t $ target graphs incurs a cost of $ \mathcal{O}(n_t \cdot L \cdot (|\mathcal{V}| + |\mathcal{E}|) \cdot d_g \cdot T) $. The total time complexity of \method{} is: $\mathcal{O}(n_s \cdot (L \cdot (|\mathcal{V}| + |\mathcal{E}|) \cdot d_g + |\mathcal{V}| \cdot (|\mathcal{V}| + |\mathcal{E}|)) + n_s^2 \cdot d_g) + \mathcal{O}(n_t \cdot L \cdot (|\mathcal{V}| + |\mathcal{E}|) \cdot d_g \cdot T)$.

\section{C. Dataset}
\label{sec:dataset}
\begin{table}[ht]
    \centering
    \caption{Statistics of the experimental datasets.}
    \begin{tabular}{lcccc}
        \toprule
        Datasets      & Graphs & Avg. Nodes & Avg. Edges & Classes \\
        \midrule
        NCI1  & 4,110   & 29.87     & 32.30     & 2       \\
        MUTAGENICITY  & 4,337   & 30.32      & 30.77      & 2       \\
        FRANKENSTEIN  & 4,337   & 16.9       & 17.88      & 2       \\
        PROTEINS  & 1,113   & 39.1      & 72.8      & 2       \\
        \midrule
        DD & 1,178 & 284.32 & 715.66 & 2 \\
        COX2 & 467 & 41.22 & 43.45& 2 \\
        COX2\_MD & 303 & 26.28 &	335.12 & 2\\
        BZR & 405 & 35.75 & 38.36 & 2 \\
        BZR\_MD & 306 & 21.30 &	225.06 & 2 \\
        \bottomrule
    \end{tabular}
    \label{tab:dataset}
\end{table}

    


\subsection{Dataset Description}

We conduct extensive experiments on four public benchmark graph datasets from TUDataset. The dataset statistics can be found in Table \ref{tab:dataset}, and
their details are shown as follows:

\begin{itemize}

\item For structure-based domain shifts:

\begin{itemize}

\item \textbf{PROTEINS.} The PROTEINS dataset~\cite{dobson2003distinguishing} contains 1,113 protein graphs. Each graph is labeled to indicate whether the corresponding protein is an enzyme. Nodes represent amino acids, and edges are constructed between amino acids within 6~\AA{} along the sequence. We divide the PROTEINS dataset into four subsets, P0, P1, P2, and P3, based on edge density, node density, and graph flux, which exhibit significant domain shifts.

\item \textbf{NCI1.} The NCI1~\cite{wale2008comparison} dataset consists of 4,100 molecular graphs, with atoms as nodes and chemical bonds as edges. Each graph is labeled based on its ability to inhibit cancer cell growth. Following the PROTEINS dataset, we divide the NCI1 dataset into four subsets, N0, N1, N2, and N3, based on edge density, node density, and graph flux. 

\item \textbf{FRANKENSTEIN.} The FRANKENSTEIN~\cite{orsini2015graph} dataset comprises 4,337 molecular graphs, with atoms as nodes and chemical bonds as edges. Each graph is labeled according to the molecule’s biological activity. Following the PROTEINS dataset, we divide the FRANKENSTEIN dataset into four subsets, F0, F1, F2, and F3, based on edge density, node density, and graph flux.

\item \textbf{MUTAGENICITY.} 
The MUTAGENICITY~\cite{kazius2005derivation} dataset includes 4,337 molecular graphs, where atoms serve as nodes and chemical bonds as edges. Labels indicate whether a compound is mutagenic, contributing to research in toxicology and chemical risk assessment. Following the PROTEINS dataset, we divide the MUTAGENICITY dataset into four subsets, M0, M1, M2, and M3, based on edge density, node density, and graph flux.

\end{itemize}

\item For feature-based domain shifts:

\begin{itemize}
\item \textbf{DD.} The DD dataset~\cite{dobson2003distinguishing} contains 1,178 graphs representing protein structures, where nodes represent amino acids and edges capture spatial or chemical proximity. DD graphs are larger and denser than PROTEINS graphs, introducing structural variations while maintaining similar label semantics.

\item \textbf{COX2.} The COX2 dataset~\cite{sutherland2003spline} contains 467 molecular graphs, while COX2\_MD includes 303 modified molecular graphs. In both datasets, nodes represent atoms and edges correspond to chemical bonds. COX2\_MD introduces structural variations to COX2 while maintaining semantic consistency, making them suitable for evaluating model robustness under domain shifts.

\item \textbf{BZR.} The BZR dataset~\cite{sutherland2003spline} consists of 405 molecular graphs. The BZR\_MD dataset contains 306 graphs with modified structures derived from BZR. Nodes correspond to atoms, and edges represent chemical bonds. BZR\_MD introduces structural variations to simulate domain shifts while maintaining consistent label semantics.

\end{itemize}

\end{itemize}

\subsection{Data processing}

In our implementation, we first process the raw graph-structured data, where each instance comprises an adjacency matrix, node features, and a graph-level label. Then, node representations are generated using available information, such as labels, attributes, or structural statistics, to ensure consistency across graphs. Furthermore, we incorporate path-based features and subgraph samples obtained through topology-aware and random sampling strategies to enhance structural representation. These components are integrated into a unified representation that captures global topology and local substructure information.

\section{D. Baselines}
\label{sec:baselines}

In this part, we introduce the details of the compared baselines as follows:

\begin{itemize}

    \item  \textbf{Graph kernel method.} We compare our \method{} with two graph kernel methods:

\begin{itemize}
    \item \textbf{WL}: The Weisfeiler-Lehman (WL) subtree method~\cite{shervashidze2011weisfeiler} iteratively refines node labels by hashing the concatenation of each node’s current label and the sorted multiset of its neighbors’ labels. This process enables efficient and expressive encoding of hierarchical structural features.
    \item \textbf{PathNN}: Path Neural Networks (PathNN) \cite{michel2023path} enhance expressiveness by explicitly modeling paths between nodes. They aggregate path-based features through attention mechanisms, capturing higher-order structural dependencies beyond immediate neighbors while preserving permutation invariance and computational efficiency.
\end{itemize}

    \item \textbf{Graph neural networks.} We compare our \method{} with four general graph neural networks:

\begin{itemize}
    \item  \textbf{GCN}: Graph Convolutional Networks (GCN)~\cite{kipf2022semi} update node representations by aggregating and transforming features from immediate neighbors. They employ a normalized adjacency matrix to ensure numerical stability and preserve local graph structure during message passing. 
    \item \textbf{GIN}: Graph Isomorphism Networks (GIN)~\cite{xu2018powerful} aggregate features from neighboring nodes using a sum operation, followed by a multi-layer perceptron. This design enables maximally expressive representations while mitigating over-smoothing through injective aggregation functions. 
    \item \textbf{GAT}: Graph Attention Networks (GAT)~\cite{velivckovic2018graph} compute node representations by assigning learnable attention weights to neighboring nodes through self-attention mechanisms. This allows for adaptive, weighted feature aggregation without relying on predefined graph structures or normalization constraints. 
    \item \textbf{GMT}: Graph Multiset Transformer (GMT) \cite{baek2021accurate} employs attention-based pooling with learnable queries to aggregate node features into graph-level representations. It decouples feature selection from structural bias, enabling flexible and expressive global information extraction. 
\end{itemize}

    \item \textbf{Label denoising methods.} We compare our \method{} with five label denoising methods:

\begin{itemize}
    \item \textbf{Co-teaching:} Co-teaching \cite{han2018co} trains two networks simultaneously, where each selects small-loss samples to teach the other, effectively filtering out noisy labels. This mutual-update strategy dynamically adjusts sample selection, enhancing robustness against severe label noise. 
    \item \textbf{RTGNN}: RTGNN~\cite{qian2023robust} introduces a noise governance framework for graph neural networks by identifying and mitigating noisy nodes through confidence estimation and adaptive neighbor selection. It further incorporates consistency regularization to ensure stable representation learning under noisy supervision.
    \item \textbf{Taylor-CE:} Taylor-CE \cite{feng2021can} enhances the robustness of cross-entropy loss to label noise by approximating it with a Taylor series expansion, which attenuates the influence of mislabeled samples through bounded gradients and smoother optimization dynamics. 
    \item \textbf{OMG:} OMG~\cite{yin2023omg} mitigates label noise in graph classification by jointly optimizing graph embeddings and label reliability. It employs an online sample reweighting mechanism that dynamically adjusts the training focus based on prediction confidence and noise estimation.
    \item \textbf{SPORT:} SPORT~\cite{yin2024sport} addresses label noise by modeling graph classification from a subgraph perspective, identifying reliable substructures through contrastive learning and sample selection. These substructures are then integrated using a noise-tolerant voting mechanism to enhance representation fidelity.
\end{itemize}

\item \textbf{Graph domain adaptation methods.} We compare our \method{} with six graph domain adaptation methods:

\begin{itemize}
    \item \textbf{DEAL:} DEAL~\cite{yin2022deal} is an unsupervised domain adaptation framework that aligns source and target domains at both the feature and prediction levels using adversarial learning. It further enhances cross-domain generalization by iteratively refining pseudo-labels through entropy minimization and consistency regularization.
    \item \textbf{CoCo:} CoCo \cite{yin2023coco} introduces a coupled contrastive framework that simultaneously aligns instance-level and class-level representations across domains to capture both fine-grained and semantic-level consistency. It leverages contrastive objectives to enhance feature discrimination and cross-domain coherence. 
    \item \textbf{SGDA:} SGDA \cite{qiao2023semi} integrates semi-supervised learning with domain adaptation by leveraging limited labeled target data to guide feature alignment between domains. It employs consistency regularization and entropy minimization to enhance representation transfer and reduce domain shift. 
    \item \textbf{A2GNN:} A2GNN \cite{liu2024rethinking} reexamines feature propagation in graph domain adaptation by introducing a structure-aware propagation strategy that mitigates noise amplification. It further incorporates domain-specific normalization to enhance stability and alignment during unsupervised adaptation. 
    \item \textbf{StruRW:} StruRW \cite{liu2023structural} introduces a structural re-weighting mechanism that dynamically adjusts the importance of nodes and edges based on their domain relevance. It enhances feature alignment by emphasizing transferable structures while suppressing domain-specific noise. 
    \item \textbf{PA-BOTH:} PA-BOTH \cite{liu2024pairwise} leverages pairwise alignment to explicitly match semantically similar node pairs across domains, enhancing structural and feature-level consistency. It further refines domain adaptation by integrating alignment signals into the representation learning process. 
\end{itemize}

\item \textbf{Label denoising domain adaptation methods.} We compare our \method{} with two label denoising domain adaptation methods:

\begin{itemize}
    \item \textbf{ALEX:} ALEX \cite{yuan2023alex} proposes a noise-robust graph transfer learning framework that addresses label noise in the source domain through adaptive label correction and structure-aware contrastive learning. It jointly refines pseudo-labels and aligns domain-invariant representations to improve cross-domain generalization under noisy supervision.
    \item \textbf{ROAD:} ROAD \cite{feng2023road} introduces a robust Unsupervised Domain Adaptation (UDA) framework that combats source label noise and domain shift by combining source sample weighting, confident target regularization, and adversarial alignment. It enhances model generalization by jointly filtering noisy labels and promoting cross-domain consistency.
\end{itemize}

\end{itemize}

For GCN, GIN, GAT, and GMT, we implement the models using PyTorch Geometric\footnote{\url{https://www.pyg.org/}}. For the other baseline methods, we utilize the official source code released by the respective authors. All models are trained using the Adam optimizer with a learning rate of $10^{-4}$, a hidden embedding dimension of 256, a weight decay of $10^{-12}$, and GNN layers of 4.

\section{E. More experimental results}

\subsection{More Performance Comparison}\label{sec:model performance}

In this part, we provide additional results for our proposed method \method{} compared with all baseline models across various datasets, as illustrated in Table~\ref{tab:proteins_idx}-\ref{tab:mutag_node}. These results consistently show that \method{} outperforms the baselines in most cases, validating the superiority of our proposed method.


\subsection{More Ablation study}\label{sec:ablation study}

To validate the effectiveness of the different components in \method{}, we conduct more experiments with four
variants on the NCI1, FRANKENSTEIN and MUTAGENICITY datasets, i.e., \method{} w/o IB, \method{} w/o EB, \method{} w/o NRL, and \method{} w/o NTR. The results are shown in Table \ref{tab:ablation_nci1} , \ref{tab:ablation_frank} and \ref{tab:ablation_mutag}. From the results, we have similar observations as summarized in Section 4.3.

\subsection{More Sensitivity Analysis}\label{sec:sensitive analysis}

\begin{figure*}[t]
    \centering
    \subfloat[NCI1]{\includegraphics[width=0.31\textwidth]{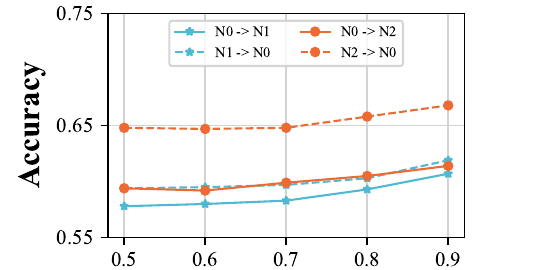}}
    \subfloat[MUTAGENICITY]{\includegraphics[width=0.31\textwidth]{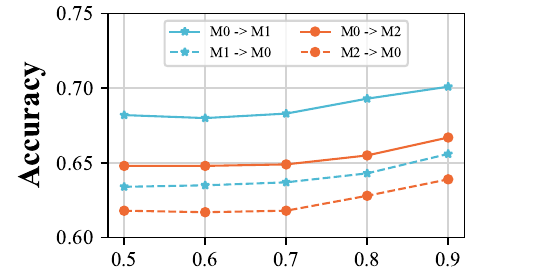}}
    \subfloat[FRANKENSTEIN]{\includegraphics[width=0.31\textwidth]{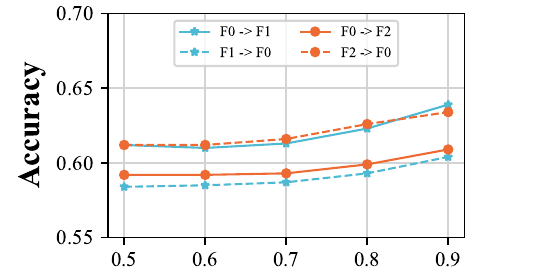}}
    \caption{Hyperparameter sensitivity analysis of threshold $\zeta$ on the NCI1, MUTAGENICITY, and FRANKENSTEIN datasets.}
    \label{fig:hyper_zeta}
    \vspace{-0.35cm}
\end{figure*}

\begin{figure*}[t]
    \centering
    \subfloat[NCI1]{\includegraphics[width=0.31\textwidth]{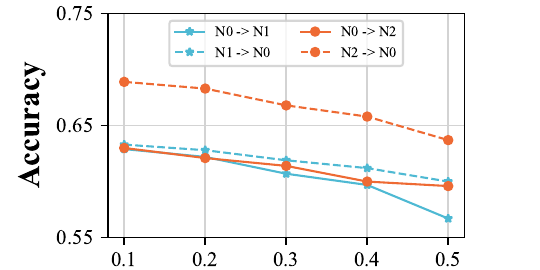}}
    \subfloat[MUTAGENICITY]{\includegraphics[width=0.31\textwidth]{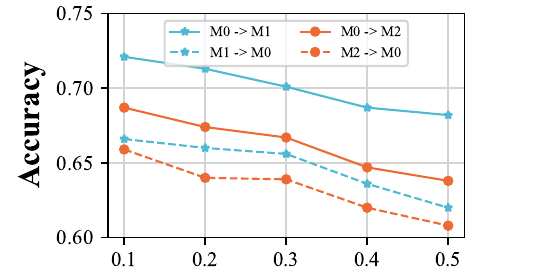}}
    \subfloat[FRANKENSTEIN]{\includegraphics[width=0.31\textwidth]{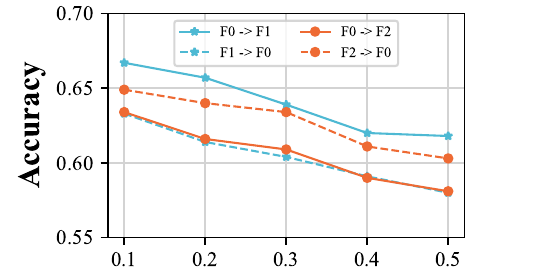}}
    \caption{Hyperparameter sensitivity analysis of noise ratio $\alpha$ on the NCI1, MUTAGENICITY, and FRANKENSTEIN datasets.}
    \label{fig:hyper_noise}
    \vspace{-0.35cm}
\end{figure*}

In this part, we provide additional results about the flexible architecturesensitivity analysis of the proposed \method{} with respect to the impact of its hyperparameters: threshold $\zeta$ and noise ratio $\alpha$ on the NCI1, FRANKENSTEIN, and MUTAGENICITY datasets. The results are illustrated in Figure \ref{fig:hyper_zeta} and  \ref{fig:hyper_noise}, where we observe trends similar to those discussed in Section 4.4.

\subsection{More Flexible Architecture}\label{sec:flexible_architecture}

\begin{figure*}[t]
    \centering
    \subfloat[NCI1]{\includegraphics[width=0.31\textwidth]{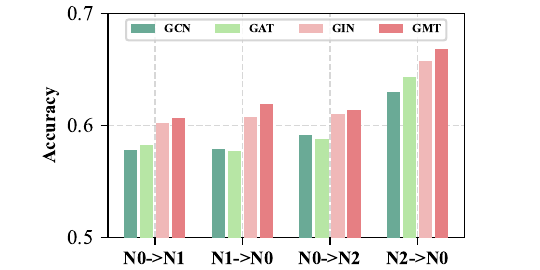}}
    \subfloat[MUTAGENICITY]{\includegraphics[width=0.31\textwidth]{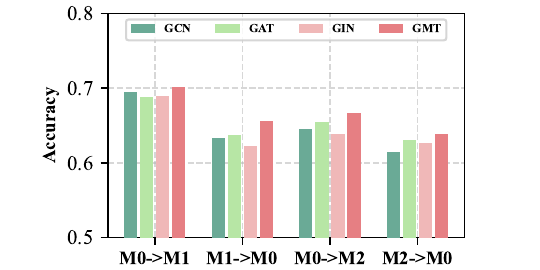}}
    \subfloat[FRANKENSTEIN]{\includegraphics[width=0.31\textwidth]{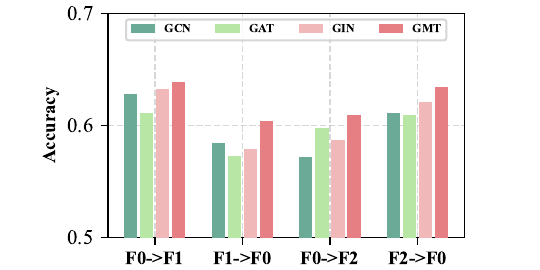}}
    \caption{The performance with different backbones for IB on the NCI1, MUTAGENICITY, and FRANKENSTEIN datasets.}
    \label{fig:different_gnn}
    \vspace{-0.35cm}
\end{figure*}

\begin{figure*}[t]
    \centering
    \subfloat[NCI1]{\includegraphics[width=0.31\textwidth]{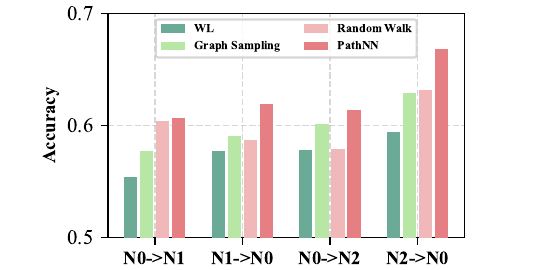}}
    \subfloat[MUTAGENICITY]{\includegraphics[width=0.31\textwidth]{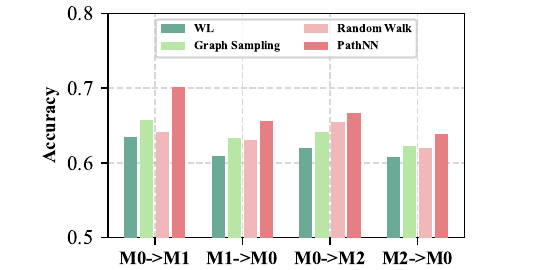}}
    \subfloat[FRANKENSTEIN]{\includegraphics[width=0.31\textwidth]{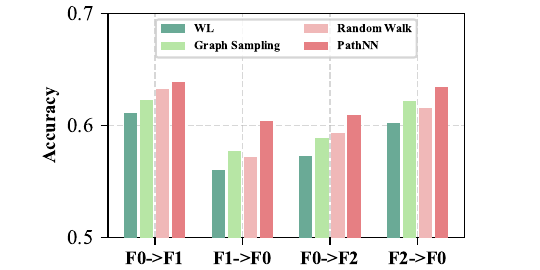}}
    \caption{The performance with different backbones for EB on the NCI1, MUTAGENICITY, and FRANKENSTEIN datasets.}
    \label{fig:different_kernal}
    \vspace{-0.35cm}
\end{figure*}

In this part, we provide additional results about the flexible architecture of the proposed \method{} on the NCI1, FRANKENSTEIN, and MUTAGENICITY datasets. The results are illustrated in Figure \ref{fig:different_gnn} and \ref{fig:different_kernal}, where we observe trends similar to those discussed in Section 4.5.

\begin{table*}[t]
\caption{The results of ablation studies on the NCI1 dataset (source → target). \textbf{Bold} results indicate the best performance per column.}
\resizebox{\textwidth}{!}{
\begin{tabular}{l|c|c|c|c|c|c|c|c|c|c|c|c}
		\toprule
		\textbf{Methods} & N0→N1 & N1→N0 & N0→N2 & N2→N0 & N0→N3 & N3→N0 & N1→N2 & N2→N1 & N1→N3 & N3→N1 & N2→N3 & N3→N2 \\ \midrule
		{\method{} w/o IB} & 51.2 & 51.3 & 49.2 & 49.9 & 49.7 & 49.5 & 49.3 & 50.2 & 52.1 & 50.0 & 53.5 & 50.3 \\
		{\method{} w/o EB} & 50.1 & 50.8 & 50.6 & 50.8 & 51.5 & 50.7 & 49.5 & 50.1 & 50.1 & 49.9 & 51.6 & 50.7 \\
		{\method{} w/o NRL} & 58.0 & 58.8 & 60.1 & 60.0 & 59.0 & 51.8 & 59.2 & 60.0 & 59.1 & 58.2 & 57.5 & 59.1 \\
		{\method{} w/o NTR} & 58.5 & 60.8 & 61.1 & 62.9 & 58.3 & 51.2 & 59.6 & 57.5 & 59.8 & 59.2 & 58.8 & 58.7 \\
		\midrule
		{\method{}} & \textbf{60.7} & \textbf{61.9} & \textbf{61.4} & \textbf{66.8} & \textbf{60.5} & \textbf{52.3} & \textbf{60.2} & \textbf{62.8} & \textbf{61.5} & \textbf{60.6} & \textbf{63.4} & \textbf{59.6} \\ \bottomrule
	\end{tabular}}
\label{tab:ablation_nci1}
\end{table*}
\begin{table}[t]
	\caption{The results of ablation studies on the FRANKENSTEIN dataset (source → target). \textbf{Bold} results indicate the best performance.}
	\label{tab:ablation_frank}
	\resizebox{\textwidth}{!}{
	\begin{tabular}{l|c|c|c|c|c|c|c|c|c|c|c|c}
		\toprule
		\textbf{Methods} & F0→F1 & F1→F0 & F0→F2 & F2→F0 & F0→F3 & F3→F0 & F1→F2 & F2→F1 & F1→F3 & F3→F1 & F2→F3 & F3→F2 \\ \midrule
		{\method{} w/o IB} & 51.4 & 49.3 & 49.5 & 50.3 & 49.5 & 49.7 & 49.7 & 50.7 & 50.4 & 50.5 & 50.4 & 50.7 \\
		{\method{} w/o EB} & 51.4 & 50.2 & 48.6 & 49.9 & 45.9 & 48.8 & 50.6 & 49.3 & 49.4 & 50.2 & 48.9 & 50.8 \\
		{\method{} w/o NRL} & 60.4 & 58.3 & 58.7 & 58.2 & 67.3 & 58.0 & 58.8 & 56.3 & 68.9 & 56.0 & 72.7 & 58.6 \\
		{\method{} w/o NTR} & 60.3 & 59.1 & 59.3 & 59.3 & 66.9 & 58.8 & 60.1 & 57.8 & 67.9 & 58.9 & 73.3 & 57.7 \\ \midrule
		{\method{}} & \textbf{63.9} & \textbf{60.4} & \textbf{60.9} & \textbf{63.4} & \textbf{68.2} & \textbf{59.9} & \textbf{60.6} & \textbf{59.7} & \textbf{72.1} & \textbf{59.6} & \textbf{74.7} & \textbf{59.4} \\ \bottomrule
	\end{tabular}}
\end{table}
\begin{table*}[t]
	\caption{The results of ablation studies on the MUTAGENICITY dataset (source → target). \textbf{Bold} results indicate the best performance.}
	\label{tab:ablation_mutag}
	\resizebox{\textwidth}{!}{
	\begin{tabular}{l|c|c|c|c|c|c|c|c|c|c|c|c}
		\toprule
		\textbf{Methods} & M0→M1 & M1→M0 & M0→M2 & M2→M0 & M0→M3 & M3→M0 & M1→M2 & M2→M1 & M1→M3 & M3→M1 & M2→M3 & M3→M2 \\ \midrule
		{\method{} w/o IB} & 52.7 & 50.3 & 50.8 & 48.5 & 50.5 & 50.9 & 51.2 & 50.7 & 50.5 & 49.5 & 48.8 & 49.4 \\
		{\method{} w/o EB} & 52.7 & 51.0 & 51.5 & 50.8 & 44.3 & 49.0 & 52.6 & 50.1 & 51.8 & 48.7 & 46.0 & 49.8 \\
		{\method{} w/o NRL} & 67.5 & 64.9 & 62.5 & 62.9 & 57.7 & 64.2 & 62.6 & 69.1 & 62.2 & 61.4 & 57.1 & 63.3 \\
		{\method{} w/o NTR} & 66.4 & 64.8 & 60.6 & 62.2 & 58.2 & 65.1 & 63.5 & 70.8 & 63.9 & 61.7 & 58.5 & 61.4 \\ \midrule
		{\method{}} & \textbf{70.1} & \textbf{65.6} & \textbf{66.7} & \textbf{63.9} & \textbf{59.7} & \textbf{65.9} & \textbf{65.8} & \textbf{72.3} & \textbf{64.2} & \textbf{63.6} & \textbf{59.1} & \textbf{65.6} \\ \bottomrule
	\end{tabular}}
\end{table*}

\begin{table*}[t]
\small
        \caption{The classification results (in \%) on the PROTEINS dataset under edge density domain shift (source $\rightarrow$ target).  P0, P1, P2, and P3 denote the sub-datasets partitioned with node density. \textbf{Bold} results indicate the best performance.}
        \label{tab:proteins_idx}
	\resizebox{1.0\textwidth}{!}{
	\begin{tabular}{l|c|c|c|c|c|c|c|c|c|c|c|c}
		\toprule                                                      
		\textbf{Methods}    & P0$\rightarrow$P1 & P1$\rightarrow$P0 & P0$\rightarrow$P2 & P2$\rightarrow$P0 & P0$\rightarrow$P3 & P3$\rightarrow$P0 & P1$\rightarrow$P2 & P2$\rightarrow$P1 & P1$\rightarrow$P3 & P3$\rightarrow$P1 & P2$\rightarrow$P3 & P3$\rightarrow$P2
 \\ \midrule
		WL  & 68.1\scriptsize{$\pm$}1.7 & 31.2\scriptsize{$\pm$}0.9 & 54.3\scriptsize{$\pm$}1.5 & 66.7\scriptsize{$\pm$}2.3 & 24.8\scriptsize{$\pm$}1.2 & 21.9\scriptsize{$\pm$}1.1 & 50.2\scriptsize{$\pm$}2.0 & 42.8\scriptsize{$\pm$}1.5 & 34.0\scriptsize{$\pm$}1.1 & 61.0\scriptsize{$\pm$}0.8 & 33.6\scriptsize{$\pm$}0.9 & 44.1\scriptsize{$\pm$}2.2 \\
        PathNN      & 68.3\scriptsize{$\pm$}1.1 & 73.0\scriptsize{$\pm$}1.2 & 55.4\scriptsize{$\pm$}0.8 & 38.5\scriptsize{$\pm$}1.0 & 25.7\scriptsize{$\pm$}0.9 & 22.9\scriptsize{$\pm$}1.1 & 40.1\scriptsize{$\pm$}0.7 & 64.0\scriptsize{$\pm$}1.4 & 34.7\scriptsize{$\pm$}1.2 & 27.9\scriptsize{$\pm$}1.3 & 67.3\scriptsize{$\pm$}0.9 & 46.9\scriptsize{$\pm$}1.0 \\
        GCN         & 67.6\scriptsize{$\pm$}1.2 & 73.7\scriptsize{$\pm$}1.1 & 56.2\scriptsize{$\pm$}0.8 & 71.8\scriptsize{$\pm$}2.0 & 24.1\scriptsize{$\pm$}1.1 & 22.8\scriptsize{$\pm$}0.7 & 52.7\scriptsize{$\pm$}1.3 & 64.3\scriptsize{$\pm$}1.5 & 27.7\scriptsize{$\pm$}1.2 & 46.0\scriptsize{$\pm$}0.9 & 30.7\scriptsize{$\pm$}1.0 & 47.3\scriptsize{$\pm$}1.4 \\
        GIN         & 62.7\scriptsize{$\pm$}1.3 & 49.1\scriptsize{$\pm$}0.8 & 50.9\scriptsize{$\pm$}1.2 & 49.7\scriptsize{$\pm$}0.9 & 25.1\scriptsize{$\pm$}1.1 & 60.3\scriptsize{$\pm$}0.7 & 45.6\scriptsize{$\pm$}1.0 & 46.1\scriptsize{$\pm$}1.2 & 66.4\scriptsize{$\pm$}1.0 & 44.6\scriptsize{$\pm$}1.4 & 43.1\scriptsize{$\pm$}0.6 & 48.8\scriptsize{$\pm$}0.9 \\
\bottomrule
	\end{tabular}}
\end{table*}

\begin{table*}[t]
\small
        \caption{The classification results on the PROTEINS dataset under node density domain shift (source $\rightarrow$ target).  P0, P1, P2 and P3 denote the sub-datasets partitioned with graph flux density. \textbf{Bold} results indicate the best performance.}
        \label{tab:proteins_node}
	\resizebox{1.0\textwidth}{!}{
	\begin{tabular}{l|c|c|c|c|c|c|c|c|c|c|c|c}
		\toprule                                                      
		\textbf{Methods}    & P0$\rightarrow$P1 & P1$\rightarrow$P0 & P0$\rightarrow$P2 & P2$\rightarrow$P0 & P0$\rightarrow$P3 & P3$\rightarrow$P0 & P1$\rightarrow$P2 & P2$\rightarrow$P1 & P1$\rightarrow$P3 & P3$\rightarrow$P1 & P2$\rightarrow$P3 & P3$\rightarrow$P2\\ \midrule
		WL  & 67.2\scriptsize{$\pm$}2.0 & 33.2\scriptsize{$\pm$}3.1 & 56.5\scriptsize{$\pm$}2.3 & 69.4\scriptsize{$\pm$}1.9 & 26.5\scriptsize{$\pm$}2.4 & 20.9\scriptsize{$\pm$}2.0 & 51.2\scriptsize{$\pm$}2.5 & 44.1\scriptsize{$\pm$}2.3 & 32.8\scriptsize{$\pm$}2.0 & 60.5\scriptsize{$\pm$}2.1 & 34.1\scriptsize{$\pm$}2.0 & 41.8\scriptsize{$\pm$}2.4 \\
        PathNN      & 69.2\scriptsize{$\pm$}2.1 & 71.2\scriptsize{$\pm$}2.2 & 55.9\scriptsize{$\pm$}2.0 & 37.7\scriptsize{$\pm$}2.1 & 26.0\scriptsize{$\pm$}2.0 & 21.8\scriptsize{$\pm$}2.2 & 41.1\scriptsize{$\pm$}2.3 & 62.8\scriptsize{$\pm$}2.0 & 33.2\scriptsize{$\pm$}2.1 & 28.2\scriptsize{$\pm$}2.0 & 66.5\scriptsize{$\pm$}2.3 & 47.0\scriptsize{$\pm$}2.1 \\
        GCN         & 68.0\scriptsize{$\pm$}2.1 & 72.0\scriptsize{$\pm$}2.0 & 56.5\scriptsize{$\pm$}2.2 & 71.7\scriptsize{$\pm$}2.3 & 25.4\scriptsize{$\pm$}2.1 & 24.1\scriptsize{$\pm$}2.0 & 54.0\scriptsize{$\pm$}2.3 & 63.6\scriptsize{$\pm$}2.2 & 28.0\scriptsize{$\pm$}2.1 & 46.0\scriptsize{$\pm$}2.0 & 31.2\scriptsize{$\pm$}2.0 & 47.4\scriptsize{$\pm$}2.1 \\
        GIN         & 61.7\scriptsize{$\pm$}2.3 & 50.2\scriptsize{$\pm$}2.4 & 52.1\scriptsize{$\pm$}2.2 & 49.9\scriptsize{$\pm$}2.1 & 43.2\scriptsize{$\pm$}2.0 & 56.5\scriptsize{$\pm$}2.3 & 44.7\scriptsize{$\pm$}2.1 & 47.0\scriptsize{$\pm$}2.2 & 58.6\scriptsize{$\pm$}2.1 & 43.1\scriptsize{$\pm$}2.0 & 42.8\scriptsize{$\pm$}2.1 & 49.2\scriptsize{$\pm$}2.2  \\
\bottomrule
	\end{tabular}}
\end{table*}
\begin{table*}[t]
\caption{The classification results (in \%) on the NCI1 dataset under graph flux density domain shift (source $\rightarrow$ target). N0, N1, N2 and N3 denote the sub-datasets partitioned with graph flux. \textbf{Bold} results indicate the best performance.}
\label{tab:nci1_flux}
	\resizebox{\textwidth}{!}{
	\begin{tabular}{l|c|c|c|c|c|c|c|c|c|c|c|c}
		\toprule
		\textbf{Methods} & N0$\rightarrow$N1 & N1$\rightarrow$N0 & N0$\rightarrow$N2 & N2$\rightarrow$N0 & N0$\rightarrow$N3 & N3$\rightarrow$N0 & N1$\rightarrow$N2 & N2$\rightarrow$N1 & N1$\rightarrow$N3 & N3$\rightarrow$N1 & N2$\rightarrow$N3 & N3$\rightarrow$N2 \\ \midrule
		{WL} & 45.3\scriptsize{$\pm$}1.2 & 55.2\scriptsize{$\pm$}3.0 & 44.3\scriptsize{$\pm$}2.4 & 35.6\scriptsize{$\pm$}1.8 & 33.5\scriptsize{$\pm$}0.9 & 30.6\scriptsize{$\pm$}2.2 & 49.2\scriptsize{$\pm$}1.5 & 53.2\scriptsize{$\pm$}2.0 & 63.3\scriptsize{$\pm$}1.6 & 51.2\scriptsize{$\pm$}3.1 & 47.5\scriptsize{$\pm$}1.0 & 47.9\scriptsize{$\pm$}2.3 \\
        {PathNN}& 53.3\scriptsize{$\pm$}1.5 & 42.1\scriptsize{$\pm$}1.3 & 54.0\scriptsize{$\pm$}0.8 & 64.2\scriptsize{$\pm$}1.9 & 42.8\scriptsize{$\pm$}2.2 & 29.3\scriptsize{$\pm$}1.1 & 50.8\scriptsize{$\pm$}2.0 & 55.7\scriptsize{$\pm$}0.9 & 60.1\scriptsize{$\pm$}2.1 & 48.3\scriptsize{$\pm$}2.3 & 58.9\scriptsize{$\pm$}1.0 & 51.5\scriptsize{$\pm$}1.7 \\
		{GCN} & 45.7\scriptsize{$\pm$}2.2 & 63.4\scriptsize{$\pm$}1.5 & 44.8\scriptsize{$\pm$}2.0 & 28.3\scriptsize{$\pm$}1.0 & 34.4\scriptsize{$\pm$}1.2 & 30.4\scriptsize{$\pm$}1.7 & 50.5\scriptsize{$\pm$}0.9 & 46.3\scriptsize{$\pm$}2.1 & 42.0\scriptsize{$\pm$}1.6 & 45.3\scriptsize{$\pm$}2.0 & 55.1\scriptsize{$\pm$}1.5 & 45.1\scriptsize{$\pm$}1.8 \\
		{GIN} & 45.7\scriptsize{$\pm$}2.4 & 63.0\scriptsize{$\pm$}2.1 & 47.2\scriptsize{$\pm$}1.6 & 34.2\scriptsize{$\pm$}2.0 & 57.0\scriptsize{$\pm$}1.0 & 28.3\scriptsize{$\pm$}1.7 & 46.2\scriptsize{$\pm$}0.9 & 47.6\scriptsize{$\pm$}2.0 & 59.3\scriptsize{$\pm$}1.5 & 49.1\scriptsize{$\pm$}2.3 & 60.8\scriptsize{$\pm$}1.9 & 49.7\scriptsize{$\pm$}0.8 \\
\bottomrule
	\end{tabular}}
\end{table*}
\begin{table*}[htbp]
\caption{The classification results (in \%) on the NCI1 dataset dataset under edge density domain shift (source $\rightarrow$ target). N0, N1, N2 and N3 denote the sub-datasets partitioned with edge density. \textbf{Bold} results indicate the best performance.}
\label{tab:nci1_idx}
	\resizebox{1.0\textwidth}{!}{
	\begin{tabular}{l|c|c|c|c|c|c|c|c|c|c|c|c}
		\toprule
		\textbf{Methods}& N0$\rightarrow$N1 & N1$\rightarrow$N0 & N0$\rightarrow$N2 & N2$\rightarrow$N0 & N0$\rightarrow$N3 & N3$\rightarrow$N0 & N1$\rightarrow$N2 & N2$\rightarrow$N1 & N1$\rightarrow$N3 & N3$\rightarrow$N1 & N2$\rightarrow$N3 & N3$\rightarrow$N2 \\ \midrule
WL & 46.7\scriptsize{$\pm$0.7} & 56.1\scriptsize{$\pm$1.8} & 45.2\scriptsize{$\pm$1.6} & 34.4\scriptsize{$\pm$1.4} & 33.3\scriptsize{$\pm$2.6} & 29.2\scriptsize{$\pm$1.5} & 48.4\scriptsize{$\pm$3.7} & 52.7\scriptsize{$\pm$1.4} & 61.7\scriptsize{$\pm$1.9} & 50.8\scriptsize{$\pm$0.7} & 46.8\scriptsize{$\pm$2.3} & 47.8\scriptsize{$\pm$2.5} \\
PathNN& 52.9\scriptsize{$\pm$2.2} & 41.4\scriptsize{$\pm$3.7} & 53.5\scriptsize{$\pm$3.0} & 63.7\scriptsize{$\pm$1.8} & 43.5\scriptsize{$\pm$1.5} & 30.1\scriptsize{$\pm$1.9} & 51.4\scriptsize{$\pm$2.6} & 55.0\scriptsize{$\pm$2.3} & 59.5\scriptsize{$\pm$1.8} & 47.6\scriptsize{$\pm$1.2} & 59.5\scriptsize{$\pm$1.4} & 52.1\scriptsize{$\pm$1.7} \\
GCN& 46.1\scriptsize{$\pm$2.8} & 64.9\scriptsize{$\pm$1.4} & 45.0\scriptsize{$\pm$1.7} & 27.9\scriptsize{$\pm$1.8} & 33.9\scriptsize{$\pm$1.3} & 29.7\scriptsize{$\pm$1.9} & 49.9\scriptsize{$\pm$0.8} & 47.1\scriptsize{$\pm$1.6} & 41.6\scriptsize{$\pm$1.9} & 44.8\scriptsize{$\pm$3.5} & 56.4\scriptsize{$\pm$3.2} & 44.5\scriptsize{$\pm$3.7} \\
GIN& 46.8\scriptsize{$\pm$2.1} & 64.4\scriptsize{$\pm$2.5} & 46.8\scriptsize{$\pm$2.2} & 33.6\scriptsize{$\pm$2.7} & 56.4\scriptsize{$\pm$1.3} & 29.7\scriptsize{$\pm$1.3} & 45.6\scriptsize{$\pm$1.7} & 47.3\scriptsize{$\pm$1.2} & 60.6\scriptsize{$\pm$0.8} & 49.9\scriptsize{$\pm$1.7} & 60.1\scriptsize{$\pm$1.3} & 50.6\scriptsize{$\pm$1.5} \\
\bottomrule
	\end{tabular}}
\end{table*}
\begin{table*}[htbp]
\caption{The classification results (in \%) on the NCI1 under node density domain shift (source $\rightarrow$ target). N0, N1, N2 and N3 denote the sub-datasets partitioned with node density. \textbf{Bold} results indicate the best performance.} \label{tab:nci1_node}
	\resizebox{\textwidth}{!}{
	\begin{tabular}{l|c|c|c|c|c|c|c|c|c|c|c|c}
		\toprule
		\textbf{Methods} & N0$\rightarrow$N1 & N1$\rightarrow$N0 & N0$\rightarrow$N2 & N2$\rightarrow$N0 & N0$\rightarrow$N3 & N3$\rightarrow$N0 & N1$\rightarrow$N2 & N2$\rightarrow$N1 & N1$\rightarrow$N3 & N3$\rightarrow$N1 & N2$\rightarrow$N3 & N3$\rightarrow$N2 \\ \midrule
		{WL} & 47.9\scriptsize{$\pm$}1.5 & 53.8\scriptsize{$\pm$}0.7 & 45.7\scriptsize{$\pm$}1.9 & 33.7\scriptsize{$\pm$}2.3 & 34.8\scriptsize{$\pm$}1.1 & 28.3\scriptsize{$\pm$}0.6 & 51.4\scriptsize{$\pm$}2.0 & 51.6\scriptsize{$\pm$}1.8 & 62.0\scriptsize{$\pm$}2.5 & 50.3\scriptsize{$\pm$}1.2 & 47.8\scriptsize{$\pm$}0.9 & 45.8\scriptsize{$\pm$}1.0 \\
        {PathNN} & 52.2\scriptsize{$\pm$}2.1 & 41.1\scriptsize{$\pm$}1.8 & 52.9\scriptsize{$\pm$}2.6 & 64.2\scriptsize{$\pm$}1.4 & 42.3\scriptsize{$\pm$}2.0 & 29.4\scriptsize{$\pm$}1.5 & 51.1\scriptsize{$\pm$}2.3 & 55.7\scriptsize{$\pm$}1.0 & 58.7\scriptsize{$\pm$}2.0 & 47.9\scriptsize{$\pm$}2.2 & 58.1\scriptsize{$\pm$}3.0 & 51.5\scriptsize{$\pm$}1.4 \\
		{GCN} & 46.6\scriptsize{$\pm$}2.1 & 63.7\scriptsize{$\pm$}1.7 & 46.8\scriptsize{$\pm$}3.1 & 26.5\scriptsize{$\pm$}0.9 & 32.2\scriptsize{$\pm$}2.4 & 31.1\scriptsize{$\pm$}2.0 & 51.2\scriptsize{$\pm$}1.6 & 48.2\scriptsize{$\pm$}1.5 & 41.0\scriptsize{$\pm$}2.2 & 45.0\scriptsize{$\pm$}2.0 & 55.8\scriptsize{$\pm$}1.5 & 45.7\scriptsize{$\pm$}3.3 \\
		{GIN} & 45.5\scriptsize{$\pm$}2.3 & 63.1\scriptsize{$\pm$}1.8 & 47.4\scriptsize{$\pm$}3.2 & 34.4\scriptsize{$\pm$}2.1 & 54.9\scriptsize{$\pm$}2.0 & 30.4\scriptsize{$\pm$}1.5 & 47.2\scriptsize{$\pm$}2.7 & 48.1\scriptsize{$\pm$}0.9 & 61.3\scriptsize{$\pm$}1.7 & 48.5\scriptsize{$\pm$}1.4 & 59.2\scriptsize{$\pm$}3.0 & 49.2\scriptsize{$\pm$}2.5 \\
\bottomrule
	\end{tabular}}
\end{table*}
\begin{table*}[t]
\small
\caption{The classification results (in \%) on the FRANKENSTEIN dataset under graph flux density domain shift (source $\rightarrow$ target). F0, F1, F2 and F3 denote the sub-datasets partitioned with graph flux density. \textbf{Bold} results indicate the best performance.}
\label{tab:frank_flux}
	\resizebox{1.0\textwidth}{!}{
	\begin{tabular}{l|c|c|c|c|c|c|c|c|c|c|c|c}
		\toprule
		\textbf{Methods} & F0$\rightarrow$F1 & F1$\rightarrow$F0 & F0$\rightarrow$F2 & F2$\rightarrow$F0 & F0$\rightarrow$F3 & F3$\rightarrow$F0 & F1$\rightarrow$F2 & F2$\rightarrow$F1 & F1$\rightarrow$F3 & F3$\rightarrow$F1 & F2$\rightarrow$F3 & F3$\rightarrow$F2 \\ \midrule
		WL & 50.4\scriptsize{$\pm$3.6} & 48.9\scriptsize{$\pm$0.8} & 49.6\scriptsize{$\pm$2.0} & 50.3\scriptsize{$\pm$2.6} & 49.7\scriptsize{$\pm$3.4} & 50.7\scriptsize{$\pm$4.2} & 48.9\scriptsize{$\pm$6.3} & 50.4\scriptsize{$\pm$2.0} & 49.4\scriptsize{$\pm$2.2} & 49.1\scriptsize{$\pm$3.3} & 50.5\scriptsize{$\pm$1.1} & 49.2\scriptsize{$\pm$1.2} \\
        PathNN & 51.9\scriptsize{$\pm$3.1} & 49.3\scriptsize{$\pm$3.3} & 49.1\scriptsize{$\pm$2.4} & 49.2\scriptsize{$\pm$2.1} & 53.0\scriptsize{$\pm$1.3} & 50.0\scriptsize{$\pm$0.8} & 49.7\scriptsize{$\pm$2.5} & 49.5\scriptsize{$\pm$1.4} & 48.9\scriptsize{$\pm$2.3} & 50.2\scriptsize{$\pm$1.1} & 49.5\scriptsize{$\pm$3.1} & 50.3\scriptsize{$\pm$1.7} \\
		GCN & 50.3\scriptsize{$\pm$1.5} & 49.9\scriptsize{$\pm$3.2} & 49.6\scriptsize{$\pm$1.0} & 50.0\scriptsize{$\pm$2.6} & 48.4\scriptsize{$\pm$1.1} & 49.4\scriptsize{$\pm$3.3} & 48.4\scriptsize{$\pm$2.5} & 49.8\scriptsize{$\pm$3.9} & 48.3\scriptsize{$\pm$7.0} & 49.2\scriptsize{$\pm$2.1} & 49.6\scriptsize{$\pm$3.7} & 50.6\scriptsize{$\pm$5.8} \\
		GIN & 50.5\scriptsize{$\pm$1.7} & 51.1\scriptsize{$\pm$0.2} & 48.4\scriptsize{$\pm$2.5} & 50.3\scriptsize{$\pm$2.3} & 48.9\scriptsize{$\pm$1.8} & 49.8\scriptsize{$\pm$2.1} & 50.9\scriptsize{$\pm$2.7} & 49.4\scriptsize{$\pm$3.4} & 55.9\scriptsize{$\pm$1.1} & 49.6\scriptsize{$\pm$3.4} & 52.1\scriptsize{$\pm$1.7} & 49.5\scriptsize{$\pm$1.5} \\
\bottomrule
	\end{tabular}}
\end{table*}
\begin{table*}[htbp]
\small
\caption{The classification results (in \%) on the FRANKENSTEIN dataset under edge density domain shift (source $\rightarrow$ target). F0, F1, F2 and F3 denote the sub-datasets partitioned with edge density. \textbf{Bold} results indicate the best performance.}
\label{tab:frank_idx}
	\resizebox{1.0\textwidth}{!}{
	\begin{tabular}{l|c|c|c|c|c|c|c|c|c|c|c|c}
		\toprule
		\textbf{Methods} & F0$\rightarrow$F1 & F1$\rightarrow$F0 & F0$\rightarrow$F2 & F2$\rightarrow$F0 & F0$\rightarrow$F3 & F3$\rightarrow$F0 & F1$\rightarrow$F2 & F2$\rightarrow$F1 & F1$\rightarrow$F3 & F3$\rightarrow$F1 & F2$\rightarrow$F3 & F3$\rightarrow$F2 \\ \midrule
		WL & 49.5\scriptsize{$\pm$1.0} & 50.7\scriptsize{$\pm$5.2} & 49.3\scriptsize{$\pm$1.7} & 50.2\scriptsize{$\pm$0.3} & 50.0\scriptsize{$\pm$2.4} & 50.1\scriptsize{$\pm$1.2} & 50.0\scriptsize{$\pm$2.4} & 49.1\scriptsize{$\pm$2.1} & 49.8\scriptsize{$\pm$3.0} & 50.5\scriptsize{$\pm$1.3} & 50.6\scriptsize{$\pm$2.3} & 49.4\scriptsize{$\pm$4.6} \\
        PathNN & 49.4\scriptsize{$\pm$3.0} & 50.1\scriptsize{$\pm$3.0} & 52.1\scriptsize{$\pm$2.3} & 50.4\scriptsize{$\pm$2.8} & 50.5\scriptsize{$\pm$11.5} & 51.5\scriptsize{$\pm$2.5} & 50.1\scriptsize{$\pm$2.0} & 50.0\scriptsize{$\pm$1.7} & 50.6\scriptsize{$\pm$2.2} & 49.8\scriptsize{$\pm$2.4} & 50.7\scriptsize{$\pm$3.7} & 50.0\scriptsize{$\pm$0.7} \\
		GCN & 49.4\scriptsize{$\pm$5.4} & 49.5\scriptsize{$\pm$3.0} & 49.1\scriptsize{$\pm$1.2} & 50.5\scriptsize{$\pm$1.4} & 50.8\scriptsize{$\pm$5.5} & 50.8\scriptsize{$\pm$1.6} & 48.2\scriptsize{$\pm$5.7} & 51.4\scriptsize{$\pm$0.5} & 51.4\scriptsize{$\pm$1.7} & 51.0\scriptsize{$\pm$2.3} & 53.1\scriptsize{$\pm$1.2} & 50.1\scriptsize{$\pm$3.1} \\
		GIN & 49.5\scriptsize{$\pm$2.5} & 50.7\scriptsize{$\pm$2.0} & 50.8\scriptsize{$\pm$9.1} & 50.0\scriptsize{$\pm$2.3} & 53.1\scriptsize{$\pm$1.2} & 50.8\scriptsize{$\pm$2.3} & 49.3\scriptsize{$\pm$1.9} & 51.5\scriptsize{$\pm$0.5} & 52.6\scriptsize{$\pm$2.6} & 51.6\scriptsize{$\pm$0.5} & 52.2\scriptsize{$\pm$2.7} & 49.3\scriptsize{$\pm$2.8} \\
\bottomrule
	\end{tabular}}
\end{table*}
\begin{table*}[htbp]
\small
\caption{The classification results (in \%) on the FRANKENSTEIN dataset under node density domain shift (source $\rightarrow$ target). F0, F1, F2 and F3 denote the sub-datasets partitioned with node density. \textbf{Bold} results indicate the best performance.}
\label{tab:frank_node}
	\resizebox{1.0\textwidth}{!}{
	\begin{tabular}{l|c|c|c|c|c|c|c|c|c|c|c|c}
		\toprule
		\textbf{Methods} & F0$\rightarrow$F1 & F1$\rightarrow$F0 & F0$\rightarrow$F2 & F2$\rightarrow$F0 & F0$\rightarrow$F3 & F3$\rightarrow$F0 & F1$\rightarrow$F2 & F2$\rightarrow$F1 & F1$\rightarrow$F3 & F3$\rightarrow$F1 & F2$\rightarrow$F3 & F3$\rightarrow$F2 \\ \midrule
		WL & 49.6\scriptsize{$\pm$0.9} & 49.6\scriptsize{$\pm$2.7} & 49.7\scriptsize{$\pm$0.8} & 48.9\scriptsize{$\pm$1.4} & 50.2\scriptsize{$\pm$1.5} & 51.1\scriptsize{$\pm$1.1} & 51.0\scriptsize{$\pm$3.1} & 49.3\scriptsize{$\pm$3.0} & 49.2\scriptsize{$\pm$3.9} & 50.1\scriptsize{$\pm$1.0} & 51.3\scriptsize{$\pm$4.2} & 50.7\scriptsize{$\pm$4.9} \\
        		PathNN & 53.1\scriptsize{$\pm$1.9} & 49.1\scriptsize{$\pm$2.0} & 50.4\scriptsize{$\pm$1.3} & 49.8\scriptsize{$\pm$1.2} & 48.8\scriptsize{$\pm$1.0} & 50.6\scriptsize{$\pm$0.9} & 50.4\scriptsize{$\pm$1.5} & 50.4\scriptsize{$\pm$1.1} & 51.2\scriptsize{$\pm$1.6} & 48.2\scriptsize{$\pm$0.4} & 49.7\scriptsize{$\pm$2.2} & 48.5\scriptsize{$\pm$2.4} \\
		GCN & 49.4\scriptsize{$\pm$1.9} & 49.9\scriptsize{$\pm$0.6} & 47.7\scriptsize{$\pm$1.7} & 50.0\scriptsize{$\pm$0.6} & 52.3\scriptsize{$\pm$2.9} & 49.5\scriptsize{$\pm$0.9} & 49.7\scriptsize{$\pm$4.4} & 51.4\scriptsize{$\pm$2.4} & 50.6\scriptsize{$\pm$2.5} & 49.5\scriptsize{$\pm$1.4} & 52.5\scriptsize{$\pm$2.5} & 48.9\scriptsize{$\pm$1.3} \\
		GIN & 46.3\scriptsize{$\pm$1.7} & 50.3\scriptsize{$\pm$0.5} & 51.4\scriptsize{$\pm$1.6} & 49.7\scriptsize{$\pm$0.3} & 51.7\scriptsize{$\pm$2.0} & 49.7\scriptsize{$\pm$0.7} & 51.8\scriptsize{$\pm$1.0} & 50.2\scriptsize{$\pm$1.5} & 48.5\scriptsize{$\pm$2.1} & 50.3\scriptsize{$\pm$1.2} & 51.9\scriptsize{$\pm$2.7} & 48.7\scriptsize{$\pm$2.2} \\
\bottomrule
	\end{tabular}}
\end{table*}
\begin{table*}[htbp]
\small
\caption{The classification results (in \%) on the MUTAGENICITY dataset under graph flux density domain shift (source $\rightarrow$ target). M0, M1, M2, and M3 denote the sub-datasets partitioned with node density. \textbf{Bold} results indicate the best performance.}
\label{tab:mutag_flux}
\resizebox{1.0\textwidth}{!}{
	\begin{tabular}{l|c|c|c|c|c|c|c|c|c|c|c|c}
		\toprule
		Methods & M0$\rightarrow$M1 & M1$\rightarrow$M0 & M0$\rightarrow$M2 & M2$\rightarrow$M0 & M0$\rightarrow$M3 & M3$\rightarrow$M0 & M1$\rightarrow$M2 & M2$\rightarrow$M1 & M1$\rightarrow$M3 & M3$\rightarrow$M1 & M2$\rightarrow$M3 & M3$\rightarrow$M2 \\ \midrule
		WL & 58.1\scriptsize{$\pm$}2.1 & 47.6\scriptsize{$\pm$}1.4 & 53.3\scriptsize{$\pm$}2.2 & 54.8\scriptsize{$\pm$}3.0 & 45.7\scriptsize{$\pm$}1.3 & 47.0\scriptsize{$\pm$}1.8 & 53.3\scriptsize{$\pm$}2.6 & 64.2\scriptsize{$\pm$}2.1 & 45.3\scriptsize{$\pm$}0.9 & 40.7\scriptsize{$\pm$}2.3 & 46.0\scriptsize{$\pm$}1.0 & 45.5\scriptsize{$\pm$}2.5 \\
        PathNN & 45.0\scriptsize{$\pm$}2.1 & 61.7\scriptsize{$\pm$}1.9 & 58.4\scriptsize{$\pm$}3.2 & 53.5\scriptsize{$\pm$}1.1 & 49.0\scriptsize{$\pm$}0.7 & 46.3\scriptsize{$\pm$}2.0 & 58.2\scriptsize{$\pm$}2.5 & 67.1\scriptsize{$\pm$}1.8 & 50.2\scriptsize{$\pm$}2.4 & 57.3\scriptsize{$\pm$}1.5 & 52.4\scriptsize{$\pm$}0.9 & 60.2\scriptsize{$\pm$}1.8 \\
		GCN & 61.1\scriptsize{$\pm$}1.8 & 52.0\scriptsize{$\pm$}2.3 & 42.9\scriptsize{$\pm$}1.1 & 47.1\scriptsize{$\pm$}0.8 & 46.7\scriptsize{$\pm$}1.2 & 56.4\scriptsize{$\pm$}1.7 & 51.4\scriptsize{$\pm$}3.0 & 33.4\scriptsize{$\pm$}1.3 & 54.9\scriptsize{$\pm$}0.6 & 37.1\scriptsize{$\pm$}2.4 & 50.6\scriptsize{$\pm$}1.5 & 58.2\scriptsize{$\pm$}2.2 \\
		GIN & 54.0\scriptsize{$\pm$}1.7 & 48.2\scriptsize{$\pm$}2.1 & 51.3\scriptsize{$\pm$}2.3 & 58.4\scriptsize{$\pm$}0.9 & 46.2\scriptsize{$\pm$}1.3 & 47.2\scriptsize{$\pm$}1.5 & 51.8\scriptsize{$\pm$}2.0 & 55.0\scriptsize{$\pm$}2.4 & 47.6\scriptsize{$\pm$}2.1 & 55.8\scriptsize{$\pm$}1.2 & 54.0\scriptsize{$\pm$}1.5 & 56.0\scriptsize{$\pm$}3.1 \\
\bottomrule
	\end{tabular}}
\end{table*}
\begin{table*}[htbp]
\small
\caption{The classification results (in \%) on the MUTAGENICITY dataset under edge density domain shift (source $\rightarrow$ target). M0, M1, M2, and M3 denote the sub-datasets partitioned with edge density. \textbf{Bold} results indicate the best performance.}
\label{tab:mutag_idx}
\resizebox{1.0\textwidth}{!}{
	\begin{tabular}{l|c|c|c|c|c|c|c|c|c|c|c|c}
		\toprule
		Methods & M0$\rightarrow$M1 & M1$\rightarrow$M0 & M0$\rightarrow$M2 & M2$\rightarrow$M0 & M0$\rightarrow$M3 & M3$\rightarrow$M0 & M1$\rightarrow$M2 & M2$\rightarrow$M1 & M1$\rightarrow$M3 & M3$\rightarrow$M1 & M2$\rightarrow$M3 & M3$\rightarrow$M2 \\ \midrule
		WL & 57.5\scriptsize{$\pm$1.2} & 48.8\scriptsize{$\pm$1.2} & 52.9\scriptsize{$\pm$1.2} & 54.2\scriptsize{$\pm$2.4} & 46.2\scriptsize{$\pm$1.8} & 47.4\scriptsize{$\pm$3.4} & 53.9\scriptsize{$\pm$2.7} & 63.9\scriptsize{$\pm$2.1} & 44.8\scriptsize{$\pm$1.7} & 41.3\scriptsize{$\pm$1.1} & 46.7\scriptsize{$\pm$3.2} & 45.7\scriptsize{$\pm$3.9} \\
        PathNN & 45.6\scriptsize{$\pm$2.3} & 60.3\scriptsize{$\pm$2.7} & 59.5\scriptsize{$\pm$2.8} & 52.8\scriptsize{$\pm$1.5} & 48.7\scriptsize{$\pm$1.9} & 47.9\scriptsize{$\pm$4.1} & 57.4\scriptsize{$\pm$0.9} & 68.2\scriptsize{$\pm$1.5} & 50.6\scriptsize{$\pm$1.6} & 58.0\scriptsize{$\pm$2.7} & 51.2\scriptsize{$\pm$1.4} & 59.6\scriptsize{$\pm$0.6} \\
		GCN & 61.1\scriptsize{$\pm$2.1} & 52.0\scriptsize{$\pm$0.9} & 43.6\scriptsize{$\pm$1.8} & 47.8\scriptsize{$\pm$1.4} & 47.1\scriptsize{$\pm$3.3} & 56.2\scriptsize{$\pm$2.6} & 51.8\scriptsize{$\pm$1.9} & 34.2\scriptsize{$\pm$2.0} & 54.2\scriptsize{$\pm$1.5} & 36.6\scriptsize{$\pm$0.7} & 50.8\scriptsize{$\pm$2.2} & 57.9\scriptsize{$\pm$1.5} \\
		GIN & 51.4\scriptsize{$\pm$0.6} & 47.7\scriptsize{$\pm$1.4} & 49.9\scriptsize{$\pm$3.6} & 57.0\scriptsize{$\pm$2.9} & 45.5\scriptsize{$\pm$1.4} & 46.5\scriptsize{$\pm$4.3} & 52.0\scriptsize{$\pm$3.5} & 53.9\scriptsize{$\pm$3.1} & 46.7\scriptsize{$\pm$1.9} & 54.7\scriptsize{$\pm$1.7} & 55.4\scriptsize{$\pm$2.1} & 55.7\scriptsize{$\pm$2.8} \\
\bottomrule
	\end{tabular}}
\end{table*}
\begin{table*}[htbp]
\small
\caption{The classification results (in \%) on the MUTAGENICITY dataset under node density domain shift (source $\rightarrow$ target). M0, M1, M2, and M3 denote the sub-datasets partitioned with node density. \textbf{Bold} results indicate the best performance.}
\label{tab:mutag_node}
\resizebox{1.0\textwidth}{!}{
	\begin{tabular}{l|c|c|c|c|c|c|c|c|c|c|c|c}
		\toprule
		Methods & M0$\rightarrow$M1 & M1$\rightarrow$M0 & M0$\rightarrow$M2 & M2$\rightarrow$M0 & M0$\rightarrow$M3 & M3$\rightarrow$M0 & M1$\rightarrow$M2 & M2$\rightarrow$M1 & M1$\rightarrow$M3 & M3$\rightarrow$M1 & M2$\rightarrow$M3 & M3$\rightarrow$M2 \\ \midrule
		WL & 56.7\scriptsize{$\pm$3.8} & 50.9\scriptsize{$\pm$4.5} & 54.4\scriptsize{$\pm$4.8} & 51.9\scriptsize{$\pm$2.1} & 46.4\scriptsize{$\pm$1.5} & 45.9\scriptsize{$\pm$1.3} & 55.1\scriptsize{$\pm$2.2} & 64.5\scriptsize{$\pm$1.8} & 42.2\scriptsize{$\pm$1.2} & 43.0\scriptsize{$\pm$3.9} & 46.6\scriptsize{$\pm$3.1} & 48.6\scriptsize{$\pm$0.8} \\
        PathNN & 43.7\scriptsize{$\pm$2.2} & 59.8\scriptsize{$\pm$4.1} & 56.9\scriptsize{$\pm$3.3} & 51.7\scriptsize{$\pm$4.2} & 50.1\scriptsize{$\pm$4.3} & 45.2\scriptsize{$\pm$2.6} & 58.0\scriptsize{$\pm$2.5} & 69.2\scriptsize{$\pm$3.0} & 51.1\scriptsize{$\pm$2.1} & 55.8\scriptsize{$\pm$2.2} & 50.3\scriptsize{$\pm$3.4} & 59.2\scriptsize{$\pm$1.6} \\
		GCN & 61.8\scriptsize{$\pm$2.0} & 51.0\scriptsize{$\pm$2.7} & 41.5\scriptsize{$\pm$3.0} & 48.2\scriptsize{$\pm$4.2} & 46.7\scriptsize{$\pm$3.9} & 57.1\scriptsize{$\pm$1.4} & 53.5\scriptsize{$\pm$4.0} & 35.5\scriptsize{$\pm$4.3} & 56.4\scriptsize{$\pm$1.5} & 34.3\scriptsize{$\pm$2.5} & 50.1\scriptsize{$\pm$1.4} & 59.7\scriptsize{$\pm$3.9} \\
		GIN & 48.9\scriptsize{$\pm$1.8} & 48.0\scriptsize{$\pm$2.3} & 48.2\scriptsize{$\pm$2.2} & 55.7\scriptsize{$\pm$3.5} & 47.5\scriptsize{$\pm$3.1} & 44.5\scriptsize{$\pm$2.3} & 53.4\scriptsize{$\pm$3.0} & 55.8\scriptsize{$\pm$2.9} & 47.2\scriptsize{$\pm$3.7} & 52.8\scriptsize{$\pm$2.8} & 56.6\scriptsize{$\pm$1.7} & 55.1\scriptsize{$\pm$3.2} \\
\bottomrule
	\end{tabular}}
\end{table*}

\end{document}